\documentclass{article}

\usepackage{arxiv}

\usepackage[utf8]{inputenc} 
\usepackage[T1]{fontenc}    
\usepackage{hyperref}       
\usepackage{url}            
\usepackage{booktabs}       
\usepackage{amsfonts}       
\usepackage{nicefrac}       
\usepackage{microtype}      
\usepackage{lipsum}		
\usepackage{tabularx}
\usepackage{multirow} 
\usepackage{adjustbox}
\usepackage{amssymb}
\usepackage{longtable}
\usepackage{booktabs}
\usepackage{pifont}
\usepackage{graphicx}
\usepackage[square,sort,comma,numbers]{natbib}
\usepackage{doi}
\usepackage{multirow}
\usepackage{float}
\usepackage{booktabs, makecell, tabularx}
\usepackage{graphicx}
\usepackage{booktabs}
\usepackage{amsmath}
\usepackage{multirow}
\usepackage{float}
\usepackage{graphicx}
\usepackage{xcolor}
\usepackage{hyperref}
\usepackage{multirow}
\usepackage{rotating}
\usepackage{pifont}

\usepackage{booktabs}
\usepackage{adjustbox}
\usepackage{tikz}
\usepackage{svg}

\title{Advancing Wheat Crop Analysis: A Survey of Deep Learning Approaches Using Hyperspectral Imaging}

\author{
\href{https://orcid.org/}{\includegraphics[scale=0.06]{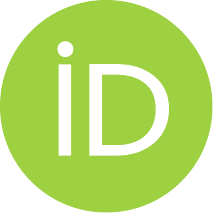}\hspace{1mm}Fadi Abdeladhim Zidi} \\
Mohamed Khider University,\\ Biskra, Algeria\\
\texttt{fadiabdeladhim.zidi@univ-biskra.dz} \\ 	
\And
\href{https://orcid.org/}{\includegraphics[scale=0.06]{orcid.pdf}\hspace{1mm}Abdelkrim Ouafi} \\
Mohamed Khider University,\\ Biskra, Algeria\\
\texttt{a.ouafi@univ-biskra.dz} \\ 
 \And
 \href{https://orcid.org/0000-0001-5077-4862}{\includegraphics[scale=0.06]{orcid.pdf}\hspace{1mm}Fares BOUGOURZI}
 \\
Junia, UMR 8520, CNRS, Centrale Lille,\\ Univerity of Polytechnique Hauts-de-France,\\ 59000 Lille, France \\
\texttt{faresbougourzi@gmail.com} \\
\And
\href{https://orcid.org/0000-0001-6581-9680}{\includegraphics[scale=0.06]{orcid.pdf}\hspace{1mm}Cosimo Distante} \\
National Research Council of Italy,\\ Monteroni ,73100, Lecce, Italy \\
\texttt{cosimo.distante@cnr.it} \\ 
\And
\href{https://orcid.org/0000-0001-7218-3799}{\includegraphics[scale=0.06]{orcid.pdf}\hspace{1mm}Abdelmalik Taleb-Ahmed} \\
Universit{\'e} Polytechnique Hauts-de-France, Université de Lille, \\CNRS, Valenciennes, 59313, Hauts-de-France, France\\
\texttt{Abdelmalik.Taleb-Ahmed@uphf.fr} \\
}

\renewcommand{\shorttitle}{}

\hypersetup{
pdftitle={},
pdfsubject={q-bio.NC, q-bio.QM},
pdfauthor={David S.~Hippocampus, Elias D.~Striatum},
pdfkeywords={First keyword, Second keyword, More},
}

\begin{document}
\maketitle

\begin{abstract}
As one of the most widely cultivated and consumed crops, wheat is essential to global food security. However, wheat production is increasingly challenged by pests, diseases, climate change, and water scarcity, threatening yields. Traditional crop monitoring methods are labor-intensive and often ineffective for early issue detection. Hyperspectral imaging (HSI) has emerged as a non-destructive and efficient technology for remote crop health assessment. However, the high dimensionality of HSI data and limited availability of labeled samples present notable challenges. In recent years, deep learning has shown great promise in addressing these challenges due to its ability to extract and analysis complex structures. Despite advancements in applying deep learning methods to HSI data for wheat crop analysis, no comprehensive survey currently exists in this field. This review addresses this gap by summarizing benchmark datasets, tracking advancements in deep learning methods, and analyzing key applications such as variety classification, disease detection, and yield estimation. It also highlights the strengths, limitations, and future opportunities in leveraging deep learning methods for HSI-based wheat crop analysis. We have listed the current state-of-the-art papers and will continue tracking updating them in the following \href{https://github.com/fadi-07/Awesome-Wheat-HSI-DeepLearning}{GitHub Repository}.
\end{abstract}


\section{Introduction}

Wheat is regarded as the most important agricultural product for global food security, contributing approximately 20\% of global dietary energy and protein intake \cite{shiferaw2013crops}. As the most widely cultivated cereal in the world \cite{acevedo2018role}, wheat grains consist of substantial amounts of carbohydrates (75\%-80\%) and protein (9\%-18\%), making them a valuable source of vitamins and minerals. However, wheat production faces numerous challenges, including pests, diseases, climate change, and the pressing need for yield optimization. These factors significantly impact crop health and overall productivity, highlighting the necessity for effective monitoring and management strategies.

To overcome challenges in wheat crop analysis, advanced remote sensing techniques like hyperspectral imaging (HSI) have gained prominence for their non-destructive, sensitive, and reliable capabilities. HSI provides detailed spectral data across diverse wavelengths, enabling precise disease detection, nutrient monitoring, and yield estimation. Modern advancements, including low-cost hyperspectral sensors like FireflEYE and HySpex VNIR, have expanded the applicability of HSI through UAVs, satellites, and ground-based systems \cite{in1,in2,in3,in4}. However, the complexity and time-consuming nature of HSI data analysis necessitate the development of automated AI-based models to ensure efficiency and accuracy while minimizing human error.

Deep Learning (DL) models for hyperspectral imaging (HSI) offer transformative advantages in wheat crop analysis, providing real-time, non-destructive insights into plant health. These techniques enable early stress and disease detection, accurate yield prediction, and optimized nutrient management, ultimately reducing costs and minimizing environmental impact \cite{ang2021big,khan2022systematic,li2022uav}. Despite significant progress in applying DL to agricultural HSI \cite{tejasree2024extensive,wang2021review,khan2022systematicc,kuswidiyanto2022plant}, existing reviews often neglect the unique challenges in wheat crop analysis, such as high spectral dimensionality, integration complexities, and limited labeled data, highlighting the need for targeted research in this area.

In the context of DL, several challenges arise when applying DL methods to wheat crop analysis using hyperspectral data. One of the most challenging aspect is the high dimensionality property of hyperspectral data, which makes it difficult for DL models to process them efficiently and often requiring considerable computational resources and specialized model architectures \cite{ abdelkrim2024hyperspectral}. Another crucial challenge is the scarcity of obtaining sufficient labeled training, which is very essential for training effective DL models. The integration of different environmental factors into DL models is challenging due to the variability in wheat crops' spectral responses to biotic and abiotic stresses \cite{kaur2024hyperspectral}. Additionally, overfitting can occur when DL models are trained on small datasets, resulting in reduced generalizability across different conditions. Addressing these challenges will require further advancements in model design and the development of more efficient training techniques \cite{sharma2024nondestructive}.

Given the lack of comprehensive surveys on wheat crop analysis using hyperspectral imaging (HSI) data, this review aims to bridge the gap by exploring the integration of HSI and deep learning (DL) techniques. We investigate how HSI technology and DL methods can be tailored to enhance wheat crop management, offering researchers and practitioners innovative solutions in this critical area. By examining the latest advancements in DL techniques applied to wheat crop analysis, this survey highlights their potential to improve crop management practices and product quality.

The existing methods are systematically categorized based on learning paradigms into: (i) supervised, (ii) semi-supervised, and (iii) unsupervised approaches. Additionally, we delve into recent advancements demonstrating significant potential in the field, including Convolutional Neural Networks (CNNs), Recurrent Neural Networks (RNNs), Transformers, Generative Adversarial Networks (GANs), Deep Belief Networks (DBNs), Stacked Autoencoders (SAEs), Diffusion models, and others. Beyond exploring these methodologies, we examine their applications, emphasizing current progress and limitations for each. To the best of our knowledge, this is the first review to focus specifically on the intersection of HSI data and DL techniques for wheat crop analysis.

The remaining sections of this review are organized as follows: Section \ref{sec:Section 2} describes the research methodology. Section \ref{sec:Section 3} provides an overview of HSI technology. Section \ref{sec:Section 4} summarizes available databases, discussing their advantages and limitations. Section \ref{sec:Section 5} details deep learning methods, while Section \ref{sec:Section 6} focuses on HSI applications in wheat crop analysis. Sections \ref{sec:Section 7} and \ref{sec:Section 8} discuss challenges and potential future research directions, respectively. Finally, Section \ref{sec:Section 9} concludes the review.

\begin{figure}[htbp]
    \centering
    \includegraphics[width=0.98\textwidth ]{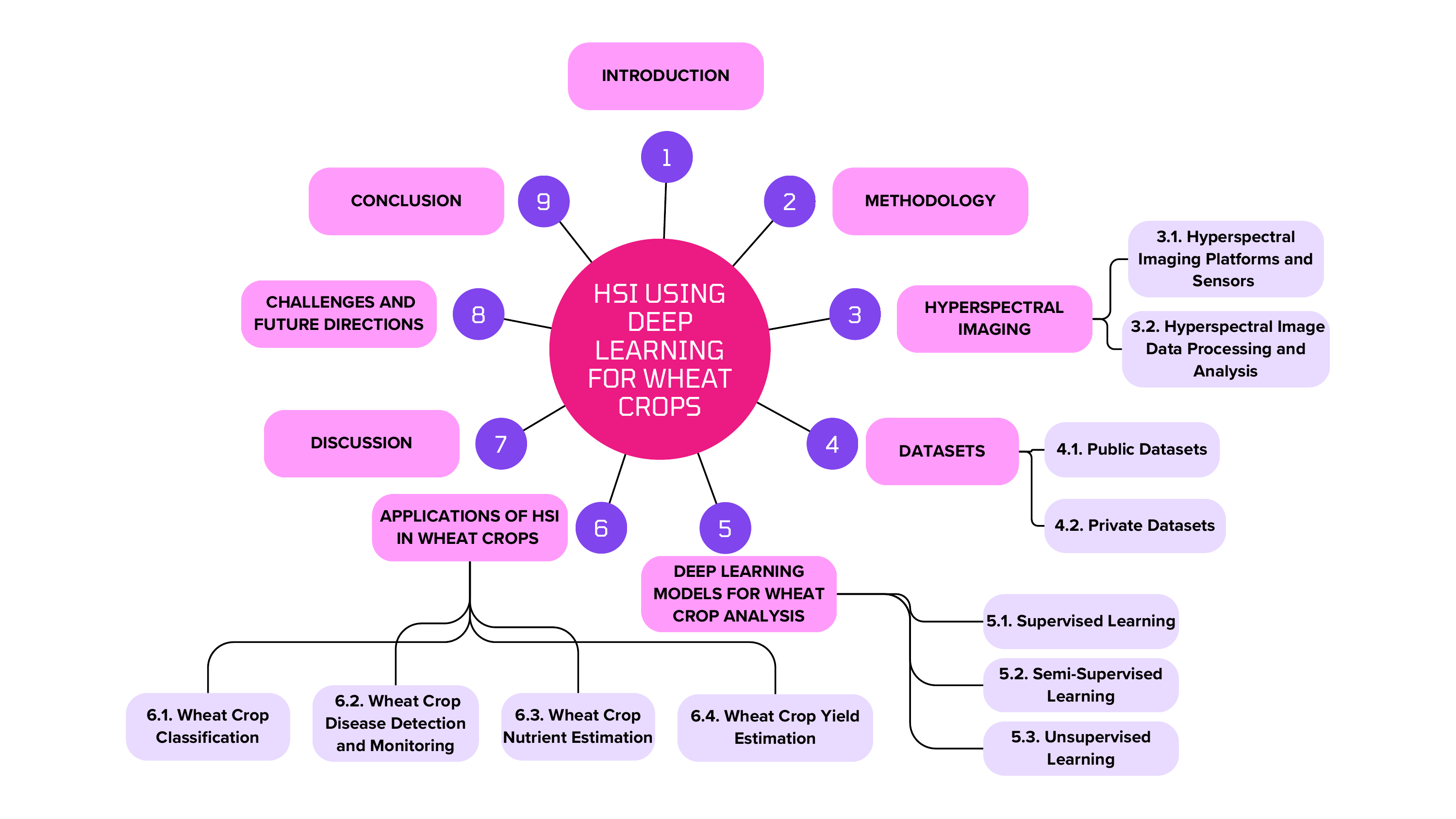}
    \caption{Visual Representation of The Survey  Structure.}
    \label{fig:Figure1}
\end{figure}

\section{Methodology}
\label{sec:Section 2}

\begin{figure}[!htbp]
    \centering
    \includegraphics[width=0.6\linewidth]{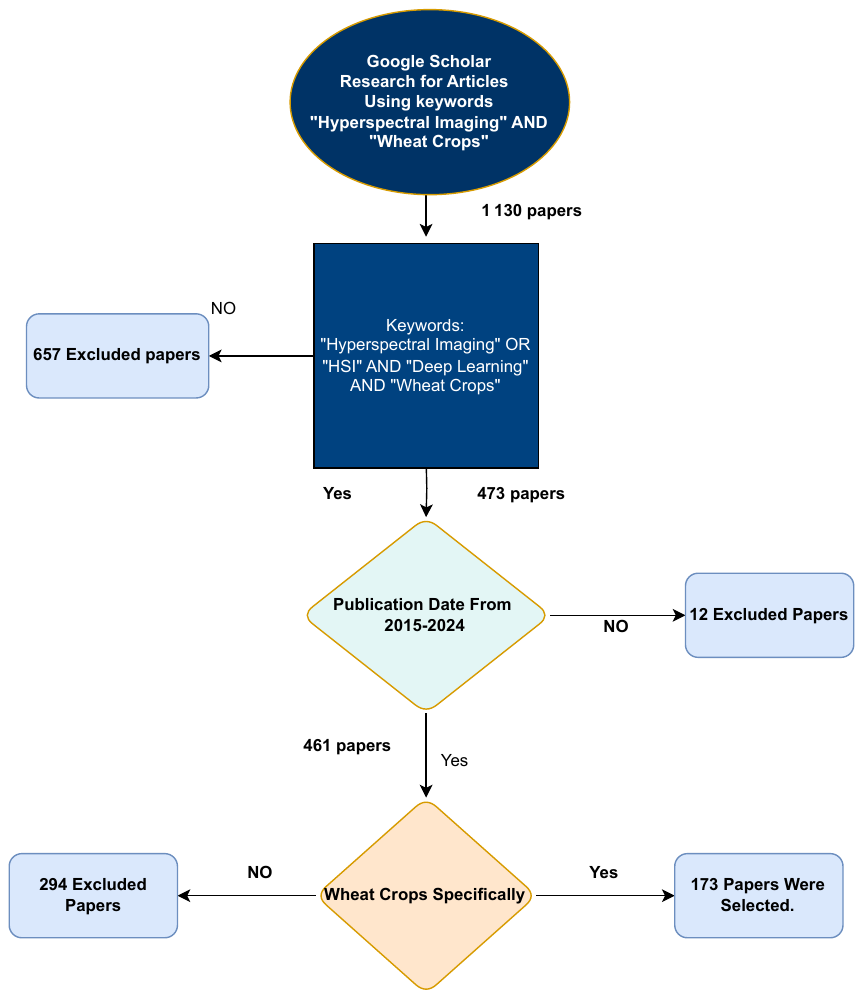}
    \caption{Overview of the Methodological Strategy for Articles Search and Selection in Conducting this Survey.}
    \label{fig:Figure2}
\end{figure}

In this section, we describe the used research strategy for collecting and analyzing the relevant research papers in this survey, as depicted in Figure \ref{fig:Figure2}. This process involves retrieving articles from several reputable scientific databases, including ScienceDirect, IEEE Xplore, SpringerLink, and MDPI. The steps followed in the screening process and review of the articles are outlined below:

\begin{enumerate}
    \item \textbf{Search in Databases}: We focused on identifying studies that utilize hyperspectral imaging and deep learning techniques for wheat crop analysis. To conduct the search, we used the following keywords: (“Hyperspectral Imaging” OR “HSI” AND “Deep Learning” AND “Wheat Crops”). The search was performed for articles published between 2015 and 2024. The search was conducted in four primary databases: ScienceDirect, IEEE Xplore, SpringerLink, and MDPI.
    
    \item \textbf{First Screening: Title, Abstract, and Methodology}: In this step, we reviewed the titles, abstracts, keywords, and methodology sections of the articles to determine if they met the inclusion criteria, which are:
    \begin{itemize}
        \item The study must be relevant to analyzing wheat crops from HSI data using deep learning techniques.  
        \item The methodology must be clear, providing sufficient details for reproduction.
        \item The results must be clearly presented.
    \end{itemize}
    
    \item \textbf{Second Screening: Full Article Review}: More details are investigated in this step, where we thoroughly reviewed the content of the articles that met the selection criteria in the first step. Articles are excluded if they do not meet the inclusion criteria, such as lacking relevant details in the methodology, not applying deep learning to hyperspectral data, or being outdated. A total of 268 articles were discarded, and 193 were retained for further analysis.
    
    \item \textbf{Analysis of Selected Articles and Information Extraction}: The final step involved analyzing and synthesizing the information from the selected articles. Key methodologies, results, and findings related to the application of deep learning models to wheat crop data were extracted. These included areas such as wheat classification, nutrient estimation, disease monitoring and detection, yield estimation, and other relevant tasks. This data was then used to identify trends, challenges, and future research opportunities in the field.
\end{enumerate}

\section{Hyperspectral Imaging}
\label{sec:Section 3}
HSI is a transformative technology designed to analyze a wide spectrum of light, contrasting with traditional RGB cameras that assign primary colors (red, green, blue) to each pixel \cite{caballero2019hyperspectral}. In HSI, the light striking each pixel is broken down into numerous spectral bands across a broad range of wavelengths. This process yields detailed information about the properties and composition of the imaged object, as different materials interact with light differently across the spectrum. By capturing narrow spectral bands over a continuous range, HSI produces spectra for all pixels in the scene \cite{example}.

As a rapidly advancing imaging modality, HSI has gained significant traction across various fields, with wheat crop agriculture being among the most notable areas of application \cite{wheat}. This section focuses on fully harnessing the potential of HSI through a detailed exploration of two key aspects: the platforms and sensors used for data acquisition, and the processing methods that enhance data quality and usability.


\begin{figure}[htbp]
    \centering
    \includegraphics[width=0.6\textwidth]{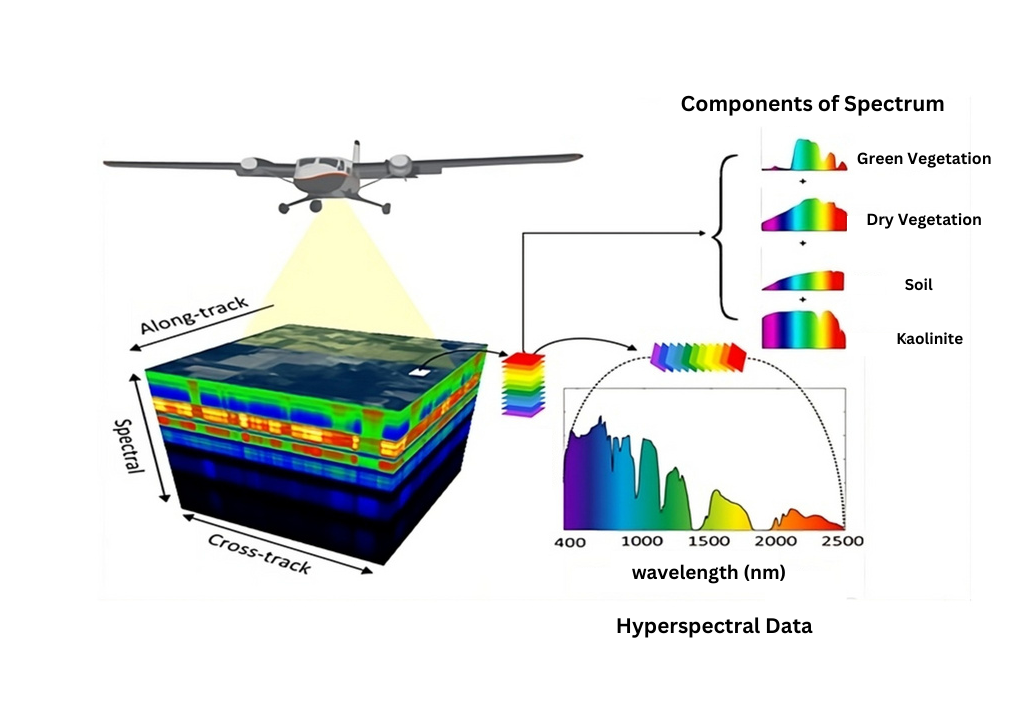}
    \caption{AVIRISng (Airborne Visible Infrared Imaging Spectrometer next generation) HSI cube of Mount Vesuvius, Italy (Image credit: NASA/JPL) \cite{pict}.}
    \label{fig:Figure5}
\end{figure}

\subsection{Hyperspectral Imaging Platforms and Sensors}

HSI platforms and sensors provide a comprehensive view of objects by utilizing a wide range of the electromagnetic spectrum \cite{adao2017hyperspectral}. These sensors operate by capturing reflected light through an imaging slit, functioning as line-scanning cameras or ‘push-broom’ sensors \cite{lu2020recent}. This mechanism allows HSI to combine spectroscopy with imaging capability, enabling the collection of detailed data that goes beyond what traditional cameras can offer \cite{lodhi2018hyperspectral}. These sensors have proven valuable in estimating the biophysical properties of landscapes, conducting thematic mapping, and analyzing surface matter and land cover \cite{vorovencii2010hyperspectral}. They can be implemented in various ways: whiskbroom (point scanning), push-broom (line scanning), and snapshot cameras.

The effectiveness of these sensors hinges on the platforms they are installed on. Hyperspectral sensors can be mounted on different types of devices, including satellites, airplanes, UAVs, and close-quarters systems. Each platform offers distinct advantages depending on the resolution and scope of the data required. For example, satellites provide broad coverage but may sacrifice some spatial detail, making them ideal for large-scale monitoring projects. UAVs, on the other hand, are increasingly popular for more localized agricultural applications. These agile platforms can gather high-resolution data at specific intervals, which is crucial for detailed analysis, such as crop disease monitoring and stress assessment.

In wheat crop applications, airplanes remain a popular choice for HSI, especially for mid-scale operations. However, the rise of UAVs in the past decade has revolutionised remote sensing, enabling precise, high-resolution imaging at much lower altitudes. UAVs are now widely considered as the most practical platforms for HSI in agriculture due to their flexibility and cost-effectiveness \cite{habib2016improving}.

To evaluate the platforms and sensors used in HSI, Table \ref{tab:Table4} outlines the most widely used sensors, detailing their specifications and operational modes, while Table \ref{tab:Table 5} showcases sensors specifically applied to wheat crop monitoring, highlighting their effectiveness in agricultural studies \cite{transon2018survey, lodhi2018hyperspectrall, jia2020overview}. The choice of platform and sensor is driven by the scale and precision needed for the study, with satellites suited for large-scale monitoring and UAVs providing high resolution for localized tasks such as nitrogen stress detection and height estimation. This approach allows for a tailored selection of sensors and platforms to meet the diverse needs of agricultural research, from broad landscape assessments to detailed crop-level analysis.


\begin{table}[htbp]
\caption{Spectral Characteristics of Remote Sensing Sensors.}
\centering
\begin{tabular}{lllll}
\toprule
\textbf{} & \textbf{Sensor} & \textbf{Spectral Range (nm)} & \textbf{Number of Spectral Bands} & \textbf{Spectral Resolution (nm)} \\
\midrule
\multirow{2}{*}{Satellite} & Hyperion & 357--2576 & 220 & 10 \\
                           & PROBA-CHRIS & 415--1050 & 19--63 & 34--17 \\

\multirow{4}{*}{Airplane}  & AVIRIS & 400--2500 & 224 & 10 \\\hline
                           & CASI & 380--1050 (CASI-1500) & 288 & $<3.5$ \\
                           & AISA & 400--970 & 244 & 3.3 \\
                           & HyMap & 440--2500 & 128 & 15 \\\hline

\multirow{2}{*}{UAV}       & Headwall Hyperspec & 400--1000 & 270--324 & 6--2.5 \\
                           & UHD185-Firefly & 450--950 & 138 & 4 \\
\bottomrule
\end{tabular}
\label{tab:sensors}
\label{tab:Table4}
\end{table}

\begin{table}[htbp]
\caption{Hyperspectral Sensors Used in Various Wheat Crop Monitoring Studies.}
\begin{tabular}{lllll}
\hline 
Platform & Sensor & Spectral Range & Application & Reference \\\hline 

\begin{tabular}{l} 
Hand-held \\
Spectrometer
\end{tabular} & ASD FieldSpec Spectroradiometer & $350-2500 \mathrm{~nm}$ & Wheat Plant Nitrogen stress & {$\cite{ranjan2012assessment}$} \\
\hline \begin{tabular}{l} 
Unmanned aerial \\
vehicle (UAV)
\end{tabular} & DJIS1000 UAV, SZDJI Technology & $450-950 \mathrm{~nm}$ & \begin{tabular}{l} 
Wheat Plant height \\
estimation
\end{tabular} & {$ \cite{tao2020estimation}$} \\
\hline Aircraft & Micro-Hyperspec VNIR model, & $400-1000 \mathrm{~nm}$ & Wheat Plant Phenotyping & {$\cite{gonzalez2015using}$} \\
\hline Airborne & \begin{tabular}{l} 
Model: A-series, Micro-Hyperspec \\
Airborne sensor, VNIR Headwall
\end{tabular} & $380-1000 \mathrm{~nm}$ & \begin{tabular}{l} 
Wheat Grain yield \\
prediction
\end{tabular} & \cite{montesinos2017predicting} \\
\hline
\end{tabular}
\label{tab:Table 5}
\end{table}

\subsection{Hyperspectral Image Data Processing and Analysis}

The data derived from HSI undergoes several processing and analysis steps, enabling the extraction of valuable scene information, including material presence and object classification \cite{chang2013hyperspectral}. This process is crucial in wheat crop applications, as it allows researchers to identify crop health, assess nutrient status, and monitor disease presence.

The initial phase of HSI data processing involves acquiring the raw data either through specialized HSI devices or pre-existing datasets. Following the acquisition, the data undergoes preprocessing to eliminate noise, correct for artifacts, and enhance overall quality \cite{sun2021spassa}. This preprocessing includes radiometric calibration to rectify sensor-specific variations, geometric correction to address distortions, and spectral calibration to align bands with known references \cite{ngadi2010hyperspectral}.

After preprocessing, data exploration is conducted to comprehend its attributes, employing techniques such as false color composites for visual representation and spectral profiles for specific pixel analysis \cite{site2}. Feature extraction is then performed to derive pertinent information, utilizing methods like spectral indices to highlight characteristics and dimensionality reduction to create a concise feature set \cite{site3,zhao2019featureexplorer}.

Finally, the decision-making phase occurs, wherein extracted features are utilized for classification and analysis \cite{xu2019superpixel}. This can involve supervised classification using labeled data, unsupervised classification via similarity-based grouping, or change detection by comparing multiple HSI samples \cite{prasad2020hyperspectral}. The recent advancements in decision making techniques will be discussed in Section \ref{sec:Section 5}.

\section{Datasets}
\label{sec:Section 4}
A new era in agricultural research and applications has dawned with the advent of HSI cameras, enabling the collection and categorization of extensive data for HSI analysis. These datasets are critical for driving progress in application of wheat crop analysis by providing the necessary data to develop and evaluate AI models \cite{hy}. In particular, deep learning methods are relied on large, diverse datasets for training and testing, well-curated hyperspectral datasets enable researchers to build models capable of accurately detecting complex patterns in wheat crops, leading to better real-world agricultural outcomes. The use of these datasets is crucial not only for improving model accuracy but also for enhancing the robustness and generalizability of DL applications across diverse environments and conditions. These datasets can be categorized as public or private dataset.


\subsection{Public Datasets}
Numerous public datasets have been collected, labeled, and released for agricultural research, forming a critical foundation for advancements in data-driven solutions. Table \ref{tab:Table1} provides a summary of publicly available datasets, including six specifically related to wheat crop analysis. These datasets support a variety of tasks, such as: (i) General agricultural datasets \cite{site5, site6}, where wheat crop is included as one class, (ii) Crop disease detection, including Soilborne Wheat Mosaic Virus \cite{site9}, Early Detection of Crown Rot \cite{site7}, and Fusarium Head Blight (FHB) detection \cite{site8}, and (iii) Salt stress phenotyping detection \cite{site10}.

\textbf{Indian Pines \cite{site5}:} The AVIRIS sensor collected this view above the IP test site in northwest Indiana. It has 224 spectral reflectance bands in the wavelength range of $0.4 $ - $ 2.5 \times 10^{-6}$ meters, and it has 145 × 145 pixels. This scene is a portion of a bigger picture. The IP scene comprises about two-thirds agricultural land and one-third forest or other natural perennial vegetation. A train line, two significant dual-lane highways, some low-density residences, other manmade buildings, and minor roadways are all present. Given that the pictures were taken in June, some of the crops—such as maize and soybeans—are only 5\% covered and are still in the early phases of development. The sixteen classes that constitute the currently accessible ground truth are not mutually exclusive, including wheat crops. Additionally, they eliminated bands [104-108], [150-163], and 220 that covered the area of water absorption, bringing the total number of bands down to 200.

\textbf{GHISA \cite{site6}:} GHISA is a comprehensive collection of hyperspectral signatures of major global crops, developed by the US Geological Survey (USGS) with global collaboration. It covers around 65\% of worldwide farmland, including crops like wheat, rice, corn, and soybeans. Hyperspectral data were collected using spaceborne, aerial, and ground-based methods. For the Conterminous United States (CONUS), GHISA provides crop data at various growth stages (GHISACONUS Version 1), using Earth Observing-1 Hyperion hyperspectral data (2008–2015) and USDA Cropland Data Layer as references for key crops like wheat and cotton.

\textbf{DRUM \cite{site10}:} This dataset includes hyperspectral pictures of four wheat lines, with one line subjected to salt (NaCl) treatment and control. The images, taken one day after salt application using a hyperspectral camera (PIKA II, Resonon), captured spectral responses in the visible and near-infrared range (400-900 nm) at high spatial resolution. Raw images were calibrated and transformed into reflectance values after removing 25 noisy spectral bands. Vegetation pixels were isolated from the background using vegetation indices and morphological operations. The dataset, primarily used for phenotyping wheat salt tolerance, is also suitable for developing algorithms for plant classification and spectral feature selection.

\textbf{Soilborne Wheat Mosaic Virus \cite{site9}:} 	
The Soilborne dataset was gathered from both laboratory and field environments. The laboratory captured HSI of wheat leaves, while additional in situ images were taken from wheat fields and disease nurseries. The dataset is intended to assist in detecting Soilborne Wheat Mosaic Virus in wheat crops. 

\begin{table}[htbp]
\caption{Public Wheat Crops Analysis datasets, where WL: Wavelength, SD: Standard Definition, NM: Not Mentioned, SWIR: Short-Wave Infrared, and VNI: Very Near Infrared.}
\centering
\resizebox{1.05\textwidth}{!}{
\begin{tabular}{lllllll}
\hline
Dataset & Year & Source & WL(nm) & Classes & SD(pixels) & Task \\
\hline
Indian Pines \cite{site5} & 1992 & NASA AVIRIS & 400-2500 & 16 & 145 × 145 & Classification of wheat class \\
GHISA \cite{site6} & 2008, 2015 & NASA Hyperion satellite & 437-2345 & 11 & 7000 & Classification of wheat class \\
Soilborne Wheat Mosaic Virus \cite{site9} & 2019 & A Nano VNIR Hyperspec & 400-1000 & 2 & NM & soilborne wheat mosaic virus
\\
Wheat HyperSpectral \cite{site8} & 2020-2021 & NM & NM & 2 & NM & Wheat Kernels for Deoxynivalenol Quantification \\
Early Detection of Crown Rot in Wheat \cite{site7} & 2021 & NM & SWIR-VNIR & 2 & NM &  Early Detection of Crown Rot in Wheat \\
DRUM \cite{site10} & 2018 & PIKA II, Resonon & 400-900 & 2 & NM & salt stress phenotyping of wheat \\
\hline
\end{tabular}%
}
\label{tab:Table1}
\end{table}

\subsection{Private Datasets}
In addition to public datasets, several recent datasets have been utilized for wheat crop analysis. However, many of these datasets are private and require specific permissions and instructions for access. These datasets primarily target two major wheat crop diseases: (i) Wheat Fusarium \cite{privi2, privi5} and (ii) Wheat Yellow Rust \cite{pri3, pri4, pri5}, as summarized in Tables \ref{tab:Table2} and \ref{tab:Table3}, respectively. The research community has focused on these two diseases because Fusarium head blight and wheat yellow rust pose significant threats to wheat production. These threats include reduced yields, lower grain quality, and contamination with mycotoxins, making them more critical compared to other wheat crop diseases. Additionally, both diseases can impair root development, limiting the plant’s ability to absorb water and nutrients and reducing leaf area \cite{figueroa2018review}.


In \cite{privi2}, E. Bauriege et al. used wheat plants (cv. ‘Taifun’) for their experiments. The plants were grown under controlled conditions in a greenhouse in pots, with 16 grains in each pot. The plants were then artificially inoculated with a mixture of pathogens of the strain Fusarium Culmorum with a spore concentration of 250,000 spores ml\textsuperscript{-1} on three successive days from the beginning of flowering. This controlled inoculation allowed us to study the development of Fusarium infection in wheat under laboratory conditions. They used HSI to analyze the plants and differentiate between spectra of diseased and healthy ear tissues in specific wavelength ranges. Consequently, the wavelength ranges most appropriate for the discrimination of head blight were determined, and the stage of grain development optimal for disease detection was also identified.

In \cite{privi3}, J. Barbedo et al. introduced a wheat disease dataset consisting of 27 HSI with 803 kernels from four wheat varieties: BRS 194, Quartzo, BRS Parrudo, and BRS 179. The dataset includes images with one-third sound kernels, one-third diseased kernels, and the rest a balanced mix. The images were captured under consistent conditions to minimize variability, aiding in accurate algorithm calibration. The HSI system used for capturing the dataset includes an objective lens, spectrometer (operating in the 528-1785 nm range with 5-7 nm spectral resolution and 256 bands), a camera, conveyor, and illumination, using push-broom acquisition to gather the data line by line.

In addition \cite{privi4}, R. Whetton et al. presented a wheat disease dataset that includes hyperspectral measurements taken at four key growth stages of wheat and barley crops. The dataset was captured using a push-broom hyperspectral imager operating within the 400–1000 nm range and features averaged spectral data aligned with visual crop assessments. RGB images were also included to enrich the dataset. The paper examined how foliar health affects yield by assigning weights to different sections of the crop canopy, providing insights into the relationship between plant health and productivity.

In \cite{privi5}, H. MA et al. introduced a dataset consisting of two hyperspectral experiments conducted in 2018 and 2019, covering the wheat-filling stages in the 350–2500 nm range. The dataset from 2018 (Exp. 1) was collected from three experimental fields in Anhui, China, during the wheat grain filling stage. In 2019 (Exp. 2), data were gathered to capture the disease incidence across all domains. The dataset includes spectral reflectance data of healthy and Fusarium head blight-infected wheat ears, which were preprocessed and standardized to reduce noise and discrepancies between the two years.

In \cite{privi1}, N. Zhang et al. utilized a dataset of HSI of diseased winter wheat spikes. The experiment, conducted in Beijing, China, involved planting winter wheat in 2.0 m × 3.0 m plots and inoculating the plants with F. graminearum during the flag leaf and anthesis stages. Six diseased wheat spikes were tested every four days until the dough stage, with thirty spikes at five different growth phases selected for analysis. The dataset includes HSI captured using hyperspectral microscope imaging (HMI) equipment.

In the dataset \cite{privi6}, D. Zhang et al. presented a wheat disease dataset with HSI data from wheat ears affected by Fusarium Head Blight (FHB) at different growth stages. The dataset included numerous narrow-band data points detailing reflectance across various wavelengths. They categorised the growth stages into late blooming, early filling, and a combination stage, and assessed FHB severity using indices like the Greenness Index (GI) and Normalized Disease Index (NDI). The data was split into training and test sets in a 3:1 ratio. Random Forest (RF) was used to identify critical wavelengths for detecting FHB, improving data processing and prediction accuracy.

In \cite{privi7}, Huang et al. introduced a wheat disease dataset consisting of in situ hyperspectral data collected from stressed wheat ears. A total of 72 samples were gathered, with 24 used as verification data and 42 as training data for developing monitoring models. This dataset was essential for training and testing models designed to identify wheat scabs. The hyperspectral data captured the physical and physiological changes induced by stress, providing detailed spectral information on the wheat ears. The research utilized this data to analyze spectral reflectance characteristics, vegetation indices, and wavelet features for detecting scab-infected wheat.

Similarly, in \cite{pri1}, W. Huang et al. introduced a wheat disease dataset consisting of multi-temporal push-broom hyperspectral imaging (PHI) obtained throughout a wheat field. The field was separated into three varieties: 98-100 (top left quadrant), Xuezao (bottom left quadrant), and Jing 411 (right half). The photos were colour-coded into five illness categories: Very Serious, Serious, Middle-range, Low-range, and None. These groups were determined using the Disease Index (DI) computed from the images. The DI was measured using 120 sample locations from the field. The correlation coefficient between PHI-derived estimations of DI and actual measured DI was high. The collection comprised spectrum reflectance measurements as well as illness incidence statistics from particular sites in the field.

In \cite{pri2}, G.Krishna et al. introduced a wheat disease dataset collected through field surveys conducted in the Indian states of Punjab and Haryana, where wheat is grown by farmers. The dataset utilized in their research was gathered in March 2013 and focused on wheat plants impacted by yellow rust disease at various severity levels. Hyperspectral reflectance data was recorded from these fields, spanning wavelengths between 350 and 2500 nm. Subsequent analysis and model building involved correlating the spectral reflectance observations with the severity of the yellow rust illness.

In another dataset \cite{pri3}, Q. Zheng et al. introduced a wheat disease dataset created using canopy spectral measurements taken with an ASD FieldSpec spectrometer, covering a spectral range of 350 nm to 2500 nm. This dataset comprises measurements collected at a height of 1.3 meters above the ground under clear skies, capturing various stages of winter wheat development. Two experiments were conducted to obtain spectral reflectance data, focusing on different growth phases and assessing the intensity of yellow rust in infected samples. The dataset played a crucial role in developing and evaluating spectral indices and models for accurate detection and tracking of Wheat Yellow Rust.

In the dataset \cite{pri4}, introduced by D. Bohnenkamp et al. HSI data from both UAVs and ground-based sources were used to identify and measure wheat yellow rust. Spectral data collected during field studies in 2018, spanning various wavelengths, were crucial for detecting disease symptoms and assessing plant reflectance. Ground-level canopy imaging was performed using a phytobike, while UAVs captured comprehensive field images from a height of 20 meters. Additionally, the dataset provides information on wavebands, spatial resolution, and ground sampling distance (GSD), which were used to evaluate annotation techniques and prediction algorithms for yellow rust identification, enhancing the detection and classification of the disease in wheat fields.

In \cite{pri5}, X. Zhang et al. presented a wheat disease dataset with HSI blocks (64 x 64 x 125 pixels) labeled as "other," "rust area," or "healthy area." A sliding-window technique was used to analyze the spectral and spatial data. This research, conducted at the China Academy of Agricultural Sciences, included four controlled wheat plots—two with yellow rust and two healthy—during a growing season with temperatures ranging from 5°C to 24°C.

In \cite{pri6}, A. Guo et al. introduced a wheat disease dataset containing spectral data from wheat leaves. This dataset comprises 288 samples, with 141 classified as healthy and 147 as sick, divided into 96 for testing and 192 for training. Their research aims to classify leaf health and identify the most relevant wavebands from a total of 406 in the HSI data, tackling challenges related to data redundancy and collinearity.

\subsection{Discussion}

While HSI datasets have been playing a crucial role in advancing wheat crop analysis research, several limitations affect their current usefulness. Many of these datasets are outdated, lacking recent data that reflects advancements in hyperspectral technology and the evolving challenges in agriculture. Furthermore, most datasets are region-specific, which limits their applicability in diverse environmental conditions. Wheat crops face unique environmental stressors depending on the region, and datasets without region-specific information may result in less accurate predictions when applied to different geographical areas.

Moreover, the majority of available datasets focus primarily on crop classification tasks and disease detection, neglecting other essential wheat crops analysis tasks such as stress monitoring and nutrient estimation, which are critical for precise wheat crop management. This absence of detailed data reduces the effectiveness of DL models in addressing real-world agricultural challenges. Consequently, researchers may find that these datasets do not fully support the development of comprehensive models for tasks like yield estimation and nutrient management.

For researchers requiring more specialized or current data, reliance on private datasets becomes necessary. However, private datasets are often costly and difficult to obtain, with access sometimes restricted to specific institutions or organizations. Acquiring these datasets can involve considerable time, effort, and financial investment, which creates significant barriers for many researchers. As a result, the ability to access high-quality, region-specific, and up-to-date data is often limited, hindering the development of advanced DL models.

These limitations highlight the ongoing need for new, non region-specific, and disease-focused datasets that better capture the complexity of wheat crop health and environmental variability. Developing more comprehensive datasets will significantly enhance the training of DL models, improving their accuracy and applicability in real-world agricultural contexts.


\begin{table}[htbp]
\caption{Private datasets for early detection of Fusarium disease in wheat: Overview of Dataset, Year, Camera Source, Wavelength, Key Bands, and Number of Classes. The meaning of the abbreviations used in this table is: WL (Wavelength).}
\centering
\begin{tabular}{llp{3.5cm}lp{3cm}l}
\hline
Dataset & Year & Source & WL(nm) & Important Bands & Classes \\ \hline
Wheat fusarium \cite{privi2} & 2011 & Specim Imspector V10E spectrograph with camera & 400–1000 & 500–533, 560–675, 682–733 & 2 \\ \hline
Wheat Fusarium \cite{privi3} & 2015 & Headwall Photonics Hyperspec Model 1003B-10151 with camera & 520–1785 & 1411 & 2 \\ \hline
Wheat Fusarium, Yellow Rust \cite{privi4} & 2018 & Gilden Photonics camera & 400–1000 & 650–700 & 2 \\ \hline
Wheat Fusarium \cite{privi5} & 2019 & ASD FieldSpec Pro spectrometer & 350–2500 & 471, 696, 841, 963, 1069, 2272 & 2 \\ \hline
Wheat Fusarium \cite{privi1} & 2019 & Surface optics SOC710VP camera & 400–1000 & 447, 539, 668, 673 & 2 \\ \hline
Wheat Fusarium \cite{privi6} & 2020 & Surface optics SOC710VP camera & 400–1000 & 560, 565, 570, 661, 663, 678 & 2 \\ \hline
Wheat Fusarium \cite{privi7} & 2020 & ASD FieldSpec Pro spectrometer & 350–2500 & 350–400, 500–600, 720–1000 & 2 \\ \hline
\end{tabular}
\label{tab:Table2}
\end{table}

\begin{table}[htbp]
\caption{Private Datasets for Early Detection of Yellow Rust Disease in Wheat. The meaning of the abbreviations used in this table is: WL (Wavelength), NM (Not Mentioned), and VNIR (Very Near Infrared).}
\begin{tabular}{llp{3.5cm}lp{2cm}l}
\hline
Dataset & Year & Source & Wavelength (nm) & Important Bands & Classes \\ \hline
Wheat Yellow Rust \cite{pri1} & 2007 & ASD FieldSpec Pro & 350–2500 & NM & 2 \\ \hline
Wheat Yellow Rust \cite{pri2} & 2014 & ASD FieldSpec Pro & 350–2500 & 428, 672, 1399 & 2 \\ \hline
Wheat Yellow Rust \cite{pri3} & 2019 & ASD FieldSpec Pro & 350–1000 & 460–720, 568–709, 725–1000 & 2 \\ \hline
Wheat Yellow Rust \cite{pri4} & 2019 & Specim ImSpector PFD V10E, Senop Oy Rikola & 400–1000, 500–900 & 594, 601, 706, 780, 797, 874, 881 & 2 \\ \hline
Wheat Yellow Rust \cite{pri5} & 2019 & Cubert S185 & 450–950 & NM & 2 \\ \hline
Wheat Yellow Rust \cite{pri6} & 2020 & Headwall Photonics VNIR, Cubert S185 & 400–1000 & 538, 598, 689, 702, 751, 895 & 2 \\ \hline
\end{tabular}
\label{tab:Table3}
\end{table}

\section{Deep learning models for Wheat Crop analysis}
\label{sec:Section 5}
Deep Learning has revolutionized hyperspectral imaging analysis, providing cutting-edge techniques to address the challenges posed by the high dimensionality and spectral redundancy inherent in HSI data. These advanced computational models enable efficient feature extraction, classification, and segmentation, making them indispensable tools for analyzing complex spectral-spatial information. Each DL method offers distinct strengths and faces specific limitations \cite{paoletti2019deep, signoroni2019deep}.

\begin{figure}[!htbp]
    \centering
    \includegraphics[width=0.82\linewidth]{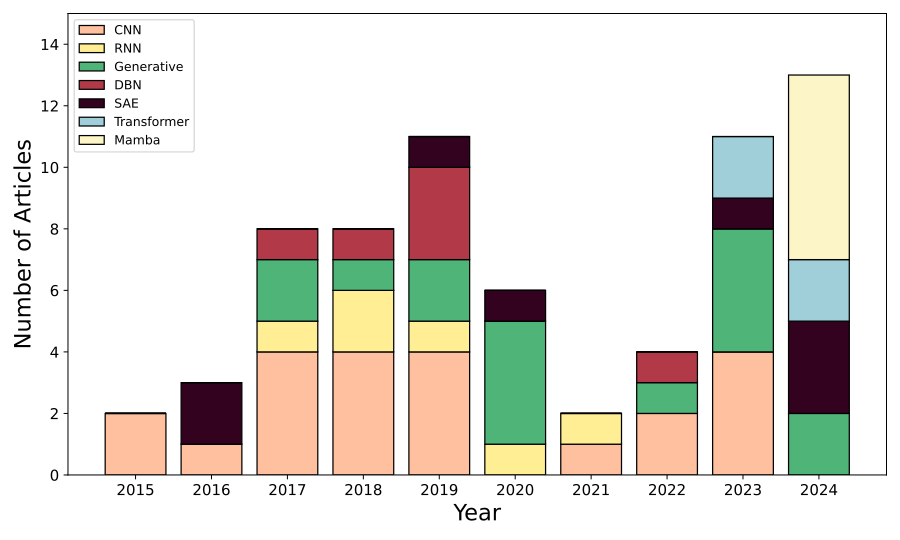}
    \caption{Number of Published Articles on DL Models by Year on Hyperspectral Data in Wheat Crops.}
    \label{fig:Figure4}
\end{figure}

Figure \ref{fig:Figure4} illustrates a timeline of DL approaches applied to wheat crop analysis using hyperspectral data. From 2015 to 2024, there has been a steady increase in publications, with notable peaks in 2017 and 2018 and a sharp surge in 2024. This trend underscores the growing interest in leveraging DL to address challenges in HSI analysis. During the early years, between 2015 and 2018, CNNs dominated the research landscape, owing to their powerful spatial feature extraction capabilities. Over time, other methods such as RNNs and DBNs were adopted to tackle changes in spectral dependencies, while generative models were employed to address data limitation challenges. More recently, advanced architectures like Transformers and Mamba models have been getting increasing interest due to  their ability to encode long-range dependencies and extracting complex relationships from hyperspectral data. The significant rise in publications in 2024 highlights the rapid advancements and innovations in DL methods tailored for HSI analysis.

\begin{figure}[!htbp]
\centering
\includegraphics[width=1\linewidth]{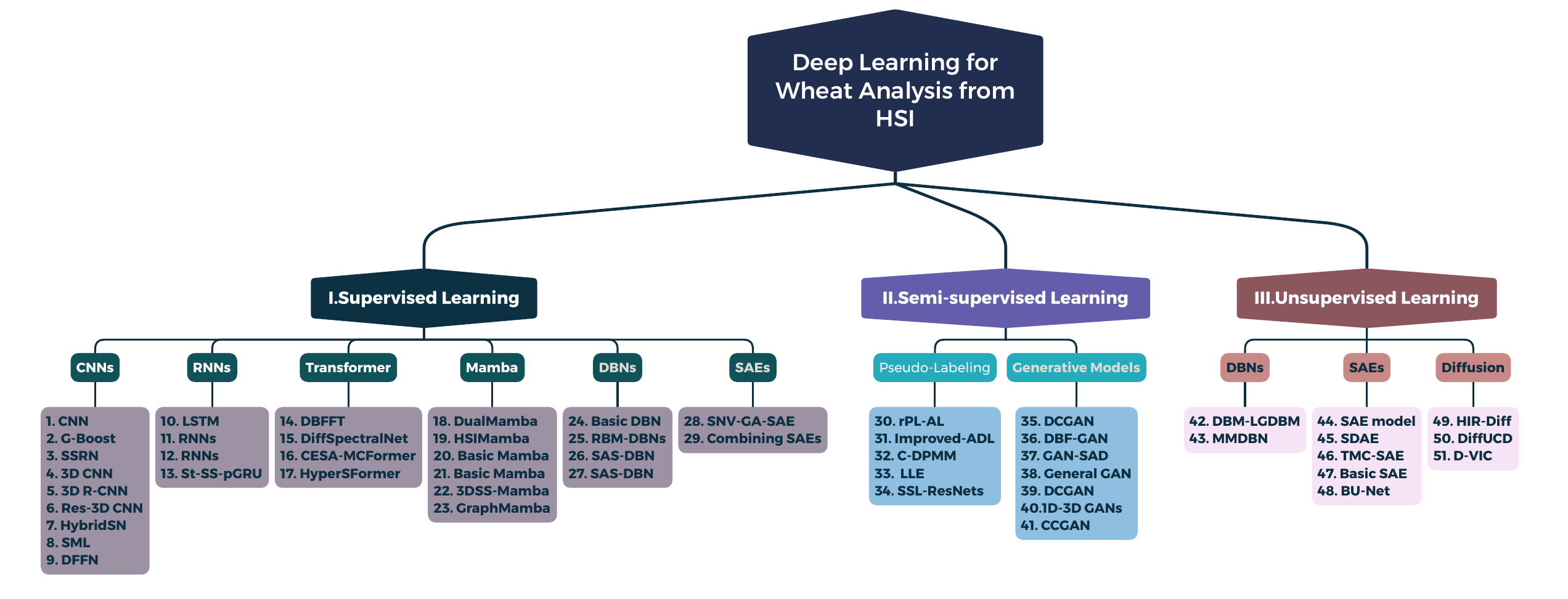}
\caption{Taxonomy for Deep Learning Methods for Wheat Crops from HSI Data:  \textbf{Supervised}:  \textbf{CNNs}: 1. \cite{sadeghi2021neural}, 2. \cite{dhakal2023machine}, 3. \cite{zhong2017spectral}, 4. \cite{chen2016deep}, 5. \cite{yang2018hyperspectral}, 6. \cite{bing2019deep}, 7. \cite{roy2019hybridsn}, 8. \cite{cnn1}, 9. \cite{song2018hyperspectral}; 
\textbf{RNNs}: 10. \cite{wang2021review}, 11. \cite{rnn1}, 12. \cite{venkatesan2019hyperspectral}, 13. \cite{luo2018shorten}; 
\textbf{Transformer}: 14. \cite{dang2023double}, 15. \cite{tran1}, 16. \cite{tran2}, 17. \cite{tran3}; 
\textbf{Mamba}: 18. \cite{sheng2024dualmamba}, 19. \cite{yang2024hsimamba}, 20. \cite{zhou2024mamba}, 21. \cite{huang2024spectral}, 22. \cite{he20243dss}, 23.\cite{yang2024graphmamba}.; 
\textbf{DBNs}: 24. \cite{dbn1}, 25. \cite{sellami2019spectra}, 26. \cite{mughees2018multiple}, 27. \cite{zhong2017learning}; 
\textbf{SAEs}: 28. \cite{xing2016stacked}, 29. \cite{bai2024two}.
\textbf{Semi-Supervised}: 
\textbf{Pseudo-Labeling}: 30. \cite{zhao2024enhancing}, 31. \cite{wang2021improved}, 32. \cite{wu2017semi}, 33. \cite{fang2018semi}, 34. \cite{zhang2020semi}
\textbf{Generative Models}: 35. \cite{gan3}, 36. \cite{he2017generative}, 37. \cite{zhan2023semisupervised}, 38. \cite{goodfellow2020generative}, 39. \cite{chen2019hyperspectral}, 40. \cite{zhu2018generative}, 41. \cite{liu2020cascade}; 
\textbf{Unsupervised}: 
\textbf{DBNs}: 42. \cite{yang2019feature}, 43. \cite{li2022manifold}. \textbf{SAEs}: 44. \cite{sae1}, 45. \cite{mughees2016efficient}, 46. \cite{afrin2024enhancing}, 47. \cite{deng2023noise}, 48. \cite{cao2024two}; \textbf{Diffusion Models}: 49. \cite{pang2024hir}, 50. \cite{zhang2023diffucd}, 51. \cite{polk2023unsupervised}}.
\label{fig:Figure6}
\end{figure}

The objective of this section is to showcase the efficacy of these DL techniques in analyzing hyperspectral data for wheat crop applications. To provide a structured analysis, the methods are categorized based on learning paradigms, which includes: supervised, semi-supervised, and unsupervised learning, as shown in Figure \ref{fig:Figure6}.

\subsection{Supervised Learning}
Supervised learning is the most widely adopted paradigm in HSI analysis, relying on labeled datasets to achieve precise and reliable predictions. This paradigm included models such as CNNs, RNNs, DBNs, SAEs, Transformers and Mamba models that employ various mechanisms to effectively process hyperspectral data \cite{10264172}.

\subsubsection{\bfseries{Convolution Neural Network}} 
\begin{figure}[htbp]
    \centering
    \includegraphics[width=1.0\textwidth]{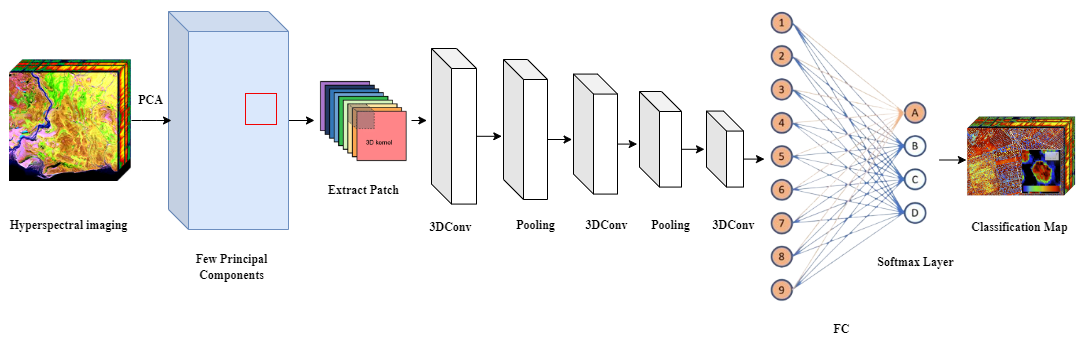}
    \caption{3D CNN Architecture Optimized for Spectral-Spatial Feature Extraction \cite{citation}. }
    \label{fig:Figure7}
\end{figure}

CNNs are particularly effective for HSI analysis because they can efficiently process high-dimensional data through convolutional layers. By leveraging 2D and 3D convolutions, CNNs excel in extracting spatial and spectral features, which are critical for analyzing the intricate spectral-spatial relationships inherent in hyperspectral datasets. Recent advancements, such as 3D CNNs, have significantly improved feature extraction and classification accuracy. Figure \ref{fig:Figure7} illustrates an example architecture of a 3D CNN, a model that uses the spectral and spatial information of HSI has been constructed. By choosing the first few primary components (bands), the model only employs a small number of informative bands. Joint spectral-spatial feature extraction is then carried out after patch extraction \cite{citation}.

Research on CNN methodologies for HSI has focused on addressing challenges related to high-dimensional data processing, efficient feature extraction, and scalable architectures. Y. Chen et al. \cite{chen2016deep} proposed a 3D convolutional architecture that processes spatial and spectral dimensions simultaneously, significantly improving feature representation, Y. Li et al. \cite{li2017spectral} advanced the use of 3D CNNs for hyperspectral classification, while X. Yang et al. \cite{yang2018hyperspectral} introduced a 3D recurrent CNN that leverages sequential relationships to enhance feature analysis. K. Dhakal et al. \cite{dhakal2023machine} proposed an ensemble CNN model, G-Boost, which achieved superior accuracy in hyperspectral data classification by integrating multiple feature learning processes.

In \cite{zhong2017spectral}, Z. Zhong et al.  developed  SSRN model, incorporating identity mapping residual blocks to improve feature learning depth and address vanishing gradient issues in hyperspectral data processing. S. K. Roy et al. \cite{roy2019hybridsn} designed the HybridSN model, combining 2D and 3D CNN layers to balance computational efficiency with robust feature extraction, enabling better spectral-spatial integration.  L. Fang et al. \cite{fang2019deep} introduced a deep hash neural network to address hyperspectral classification challenges, and W. Song et al. \cite{song2018hyperspectral} proposed a feature fusion network to enhance robustness in high-dimensional data analysis. Recent advances in the field highlight deeper architectures, lightweight designs, and the integration of techniques such as residual learning and ensemble methods. Notably, H. Lee and H. Kwon \cite{lee2017going} introduced a 12-layer 3D CNN with residual connections, enabling effective hierarchical feature learning.

\subsubsection{\bfseries{Recurrent Neural Networks}}

RNNs are designed to handle sequential data by utilizing feedback connections to capture dependencies between sequence elements, making them well-suited for tasks like HSI analysis. However, traditional RNNs struggle with long-range dependencies due to issues such as gradient vanishing and explosion \cite{chung2014empirical}. To address these limitations, Long Short-Term Memory (LSTM) networks were developed, incorporating gating mechanisms to effectively capture long-range dependencies by maintaining information over time \cite{graves2012long}. Gated Recurrent Units (GRUs) further simplify LSTMs by combining forget and input gates into a single update gate, offering similar benefits with reduced computational complexity \cite{chung2014empirical}.

RNN-based models have been widely adopted for HSI analysis. L. Mou et al. \cite{mou2017deep} used RNNs to treat hyperspectral pixels as sequential data, enabling effective information categorization and tackling the high dimensionality of HSI data. RNN models have also emerged as powerful tools for HSI analysis with R. Venkatesan and S. Prabu \cite{venkatesan2019hyperspectral} used RNNs to treat spectral data as sequences of information, leveraging spectral dependencies effectively, and employing specialized activation functions to stabilize training and enhance performance. Ch. Wang et al. \cite{wang2021review} introduced LSTM variants, including Bi-CLSTM, for spectral-spatial integration in HSI, eliminating the need for sliding-window segmentation and achieving a 1.5\% improvement in classification accuracy over 3D-CNNs in Indian Pines dataset. In \cite{rnn3}, E. Ndikumana et al. demonstrate the potential of RNNs in remote sensing tasks, such as agricultural land cover mapping using Sentinel-1 radar data, achieving better categorization outcomes compared to traditional models by explicitly leveraging temporal dependencies in multitemporal datasets.

L. Salmela et al. \cite{salmela2021predicting} introduced an LSTM-based RNN to model ultrafast nonlinear dynamics in optical fibers, effectively bypassing traditional numerical methods like the NLSE. This approach achieves faster and more accurate modeling by taking advantage of the RNN's ability to process sequential data and retain long-term dependencies via internal memory. However, the reliance of the method on simulation-generated training data raises concerns about its applicability to real-world scenarios.

H. Luo et al. \cite{luo2018shorten} introduced a shorten spatial spectral RNN with Parallel-GRU for HSI classification, which combines 3D convolutional layers and a novel Parallel-GRU architecture to efficiently capture both spatial and spectral features. The model leverages a shorten RNN structure to reduce computational complexity and improve training efficiency, while the Parallel-GRU architecture enhances robustness and performance by aggregating outputs from multiple GRU units. However, the model's reliance on carefully designed subgraphs and convolutional kernels may limit its generalization to diverse HSI datasets with varying spatial and spectral characteristics.

\subsubsection{\bfseries{Deep Belief Network}}

DBNs are hierarchical generative models comprising stacked Restricted Boltzmann Machines (RBMs) that learn layered representations of data through unsupervised training but can also be adapted to labeled datasets \cite{hinton2006fast}. Their ability to extract spectral-spatial features and reduce dimensionality makes them well-suited for HSI classification. For example, Mughees and Tao \cite{mughees2018multiple} developed a Spectral-Adaptive Segmented DBN, which partitions spectral bands into subgroups, processes them with localized DBNs, and integrates spatial features via hypersegmentation to improve accuracy and computational efficiency. Similarly, Ch. Li et al. \cite{dbn1} proposed a novel DBN-based method that employs multivariate optical sensors and a two-phase training strategy. The method features unsupervised pretraining using RBMs to extract latent features and supervised fine-tuning to optimize classification objectives with labeled data. This separation of training phases enhances feature extraction and predictive accuracy, effectively addressing the high-dimensionality of hyperspectral datasets. DBN-based methodologies continue to demonstrate significant potential in improving classification accuracy and computational efficiency. Beyond these works, DBN-based methodologies show significant potential in tasks such as identifying subtle spectral variations (e.g., wheat disease classification) and operating with limited labeled samples. Figure \ref{fig:Figure11} illustrates a DBN architecture by P. Zhong et al. \cite{zhong2017learning}, where gray nodes denote input features, and yellow/red nodes represent binary latent variables (0 and 1), highlighting the model’s capacity to capture shared spectral-spatial patterns.

 A hybrid method proposed by A. Sellami and IR. Farah \cite{sellami2019spectra} introduces a graph-based supervised learning framework for HSI classification. This approach constructs a spectro-spatial graph, where nodes correspond to pixel spectral vectors, and edges capture spatial relationships between neighboring pixels. To preserve the local manifold structure of the data, the method leverages a Restricted Boltzmann Machine for feature extraction. Classification is subsequently performed using a Deep Belief Network combined with Logistic Regression, achieving notable accuracy in HSI tasks. The study highlights the effectiveness of graph-based methods in addressing the complex challenges of hyperspectral data analysis.

\subsubsection{\bfseries{Stacked Autoencoder}}

\begin{figure}[htbp]
    \centering
   \includegraphics[width=0.8\textwidth]{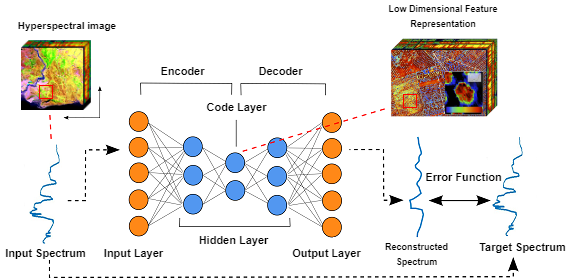}
    \caption{Optimized Stacked Autoencoder Architecture for HSI \cite{windrim2019unsupervised}.}
    \label{fig:Figure12}
\end{figure}

SAEs  have been effectively applied to supervised learning tasks in HSI analysis. By hierarchically extracting features, SAEs reduce the dimensionality of HSI data, capturing both spectral and spatial information essential for accurate classification \cite{wan2017stacked}. Figure \ref{fig:Figure12} depicts an autoencoder architecture for hyperspectral data. From which, each spectrum is the input and reconstruction target, with the code layer creating a compact feature representation. The trained encoder maps spectra to a lower-dimensional feature space \cite{windrim2019unsupervised}.
In supervised learning, SAEs are typically pre-trained in an unsupervised manner to learn robust feature representations from unlabeled data. Subsequently, these representations are fine-tuned using labeled data to optimize performance for specific tasks, such as land cover classification or object detection. This approach enhances the model's ability to generalize from limited labeled samples, a common scenario in HSI analysis \cite{bai2024two}.

Recent advancements in SAEs for HSI have introduced diverse architectures, including standard SAEs, convolutional SAEs, and hybrid designs. While basic SAEs focus on dimensionality reduction, advanced models integrate spectral and spatial feature extraction and hybrid approaches often combine SAEs with techniques like genetic algorithms or supervised fine-tuning to optimize classification performance on complex HSI datasets. K. Liang et al. \cite{sae1} introduced an SAE model combined with a genetic algorithm (GA) for wavelength selection, achieving efficient detection of deoxynivalenol (DON) levels in wheat. Their SNV-GA-SAE model excelled for wheat flour samples, while the MSC-GA-SAE model performed well for wheat kernel classification. Y. Bai et al. \cite{bai2024two} proposed a Two-stage Multi-dimensional Convolutional Stacked Autoencoder framework, which employs spectral feature extraction through dimensionality reduction in the first stage and spatial feature extraction to capture local neighborhood patterns in the second stage. Both stages are fine-tuned in a supervised manner, optimizing classification performance while addressing data scarcity challenges. Ch. Xing et al. \cite{xing2016stacked} utilized Stacked Denoising Autoencoders to reconstruct inputs corrupted with noise during unsupervised pretraining, effectively capturing underlying data structures and enhancing classification accuracy through supervised fine-tuning for HSI tasks. The evolution of SAEs continues through hybrid architectures and advanced pre-training strategies, such as combining genetic algorithms with autoencoder models, which have demonstrated significant improvements in feature selection and classification accuracy \cite{singh2022enhanced}.

\subsubsection{\bfseries{Transformer}}
\label{subsec:Transformer}
\begin{figure}[htbp]
    \centering
    \includegraphics[width=0.90\linewidth]{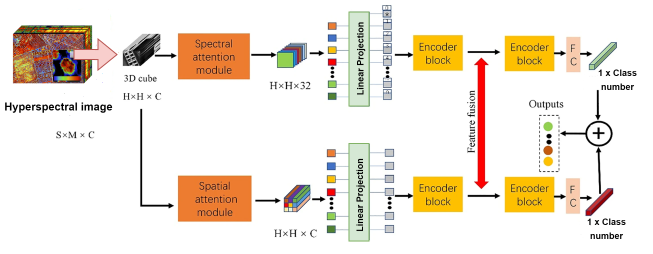}
    \caption{The overall structure of the proposed model using the transformer architecture \cite{dang2023double}.}
    \label{fig:Figure9}
\end{figure}

Transformers were initially introduced for Natural Language Processing (NLP) tasks, with self-attention as their foundational building block \cite{vaswani2017attention}. This mechanism excels in capturing long-range dependencies within input sequences by effectively weighting the relationships between elements, irrespective of their distance. The architecture's success in NLP has led to its adaptation across various domains, including HSI. In the context of HSI, Transformers demonstrate remarkable capabilities in extracting complex spectral-spatial features, offering significant advantages over traditional methods \cite{he2021spatial}. Figure \ref{fig:Figure9} shows an example Transformer model proposed by L. Dang et al. \cite{dang2023double}, which incorporates spectral and spatial branches enhanced with attention modules and Transformer encoder blocks. A feature fusion layer, inspired by CrossViT34, seamlessly integrates features from both branches to enable robust and efficient processing of spectral and spatial information.

Several Transformer-based models have been proposed for HSI analysis, showcasing their adaptability and effectiveness in capturing spectral-spatial relationships. N. Sigger et al. \cite{tran1} proposed DiffSpectralNet integrates diffusion models and transformers to enhance HSI classification by effectively capturing complex spectral-spatial relationships. Using unsupervised feature extraction via denoising diffusion probabilistic models and supervised classification with a hierarchical transformer classifier, it addresses the challenges of high-dimensional HSI data. Sh. Liu et al. \cite{tran2} proposed the CESA-MCFormer architecture, which incorporates a Center Enhanced Spatial Attention module and Morphological Convolution to minimize redundancy in spectral and spatial data while enhancing central pixel focus and maintaining global spatial information. Unlike traditional methods reliant on dimensionality reduction techniques such as PCA, this low-parameter model delivers robust classification across multiple datasets. J. Xie et al. \cite{tran3} introduced the HyperSFormer architecture, an advanced version of the SegFormer architecture, which leverages an improved Swin Transformer for encoding global spectral-spatial features and a decoder for precise classification. Key features of this model include the Hyper Patch Embedding module for effective feature extraction, the Transpose Padding Upsample module for accurate end-to-end classification, the Adaptive Min Log Sampling strategy, and using a combination of dice and focal losses to address sample scarcity and imbalance, thus improving performance on challenging HSI datasets.

\subsubsection{\bfseries{Mamba}}
Recently, this method was introduced as a replacement transformer for long-range dependencies encoding with lower computational cost, building on the Structured State Space (S4) model \cite{gu2021efficiently}. Its distinctive selection technique enables computational efficiency, flexibility with irregular data, and applicability in domains such as text analysis, audio processing, and real-time translation by adjusting the parameters of the structured state-space model according to the input \cite{gu2023mamba}. Mamba has also been widely investigated for HSI classification, where its lightweight architectures and innovative techniques effectively capture spectral and spatial relationships. These Mamba-based models have demonstrated high accuracy and computational efficiency in various datasets.

In \cite{sheng2024dualmamba},  J. Sheng et al. introduced DualMamba network, a lightweight framework that integrates global and local feature extraction, achieving high overall accuracy on multiple datasets while significantly reducing computational costs.  Yang et al. \cite{yang2024hsimamba} proposed HSIMamba model, which incorporates data augmentation strategies to improve robustness and performance, achieving an OA of 0.899\% and a Kappa coefficient of 0.8857 on the Indian pines dataset, thus outperforming traditional architectures in managing long sequences of hyperspectral data. W. Zhou et al. \cite{zhou2024mamba} emphasized a Mamba-based framework that integrates spectral and spatial data, leveraging the computational efficiency of Mamba modules to achieve high accuracy with minimal resource demands. A. Yang et al. \cite{yang2024graphmamba} introduced the GraphMamba framework to address challenges such as spectral redundancy and complex spatial relationships, incorporating graph structure learning to deliver superior classification performance across diverse datasets. Y. He et al. \cite{he20243dss} introduced 3D-Spectral-Spatial Mamba model, which employs a novel sequence modeling approach based on the State Space Model (SSM). This framework includes a Spectral-Spatial Token Generation module and a 3DSS mechanism to capture global spectral interactions and extract discriminative spectral-spatial features, achieving linear computational complexity while maintaining high classification performance. L. Huang et al. \cite{huang2024spectral} investigated the performance of Mamba-based models in various HSI applications such as wheat. Collectively, these models address practical challenges in hyperspectral data analysis by improving classification accuracy, computational efficiency, and robustness, exemplifying progress in both performance and scalability for HSI tasks.

\subsection{Semi-Supervised Learning}
Semi-supervised learning (SSL) methods offer a powerful approach by exploiting both labeled and unlabeled data to improve classification performance. These methods are especially valuable in HSI, where acquiring labeled samples is both costly and time-consuming. The used semi-supervised techniques for wheat crops analysis can be categorized into two main categories: pseudo-labeling and generative models.

\subsubsection{\bfseries{Pseudo-Labeling}}
Pseudo-labeling techniques play a vital role in HSI by enabling the effective utilization of unlabeled data through the iterative generation and refinement of labels based on model predictions. This approach is particularly beneficial in HSI, where obtaining labeled data can be both costly and time-consuming, allowing for more accurate and robust analysis of complex spectral information\cite{kotzagiannidis2021semi}. J. Zhao et al. \cite{zhao2024enhancing} introduced a semi-supervised framework for HSI classification that combines pseudo-labeling and active learning. A progressive pseudo-label selection strategy leverages spatial–spectral consistency to iteratively refine labels by linking high-confidence samples within semantically connected regions. The model employs an architecture, integrating 3D-CNNs for spectral-spatial feature extraction and vision transformers for global contextual understanding. Active learning extends this approach by selecting high-value regions or samples for expert annotation based on temporal confidence differences across rounds. This iterative framework is optimized with confidence thresholds and a hybrid classification backbone to maximize the use of limited labeled samples and abundant unlabeled data.
 Expanding on the potential of active learning, Q. Wang et al. \cite{wang2021improved} proposed an improved semi-supervised classification framework for HSI that combines active deep learning with a random multi-graph algorithm. It replaces expert labeling with an anchor graph approach to generate pseudo-labels, expanding the training set iteratively. The framework begins with a limited labeled dataset to initialize a CNN, which predicts category probabilities for unlabeled data. Informative samples are selected using AL strategies like entropy and best-versus-second-best (BVSB) metrics, followed by pseudo-labeling through the random multi-graph algorithm. Newly labeled samples are iteratively added to refine the CNN model. Key contributions include eliminating manual labeling and integrating anchor graphs for efficient pseudo-labeling, resulting in state-of-the-art accuracy across multiple datasets. However, the method's performance heavily relies on the quality of initial labels and the robustness of pseudo-labeling under noisy conditions.

Complementing these active learning strategies, H. Wu et al. \cite{wu2017semi} utilized semi-supervised learning to incorporate pseudo-labels derived from clustering algorithms, which enhance feature learning and mitigate overfitting. This method leverages pseudo-labels to pre-train networks, enabling better initialization and faster convergence during fine-tuning. It also effectively handles unknown background classes, making the approach robust for real-world applications. Similarly, B. Fang et al. \cite{fang2018semi} employed a co-training framework where spectral and spatial ResNets were trained independently, with each network refining the pseudo-labels generated by the other. Their model integrates 1D CNNs for spectral features and PCA-reduced 2D CNNs for spatial features, ensuring robustness through consistency regularization.

In \cite{chen2024prototype}, R. Chen et al. proposed Prototype-Based Pseudo-Label Refinement (PPLR) for semi-supervised HSI classification, addressing noisy pseudo-labels through adaptive thresholds and class prototypes. By dynamically adjusting confidence thresholds for pseudo-label generation and refining labels via spectral-spatial prototypes derived from labeled and unlabeled data, PPLR filters misclassified samples while assigning reliability weights to enhance robustness. The integration of a center loss to minimize intraclass feature distances further improved discrimination, achieving state-of-the-art accuracies of 82.11\% on the Indian Pines dataset with minimal labeled samples.

\subsubsection{\bfseries{Generative Models}}
generative models learn underlying patterns or distributions of data to generate new similar data. These models are particularly impactful in semi-supervised learning, where they leverage both labeled and unlabeled data, making them efficient for addressing data scarcity in HSI \cite{he2017generative}.
Building on the exploration of generative models, such as GANs, in HSI classification, several advanced strategies have emerged to address challenges like limited labeled data, noisy inputs, and the need for better data utilization. Combining semi-supervised learning with sophisticated feature extraction methods aims to enhance classification accuracy and improve the overall performance of HSI models \cite{tao2020semisupervised}. 

\begin{figure}[htbp]
    \centering
     \includegraphics[width=0.95\textwidth ]{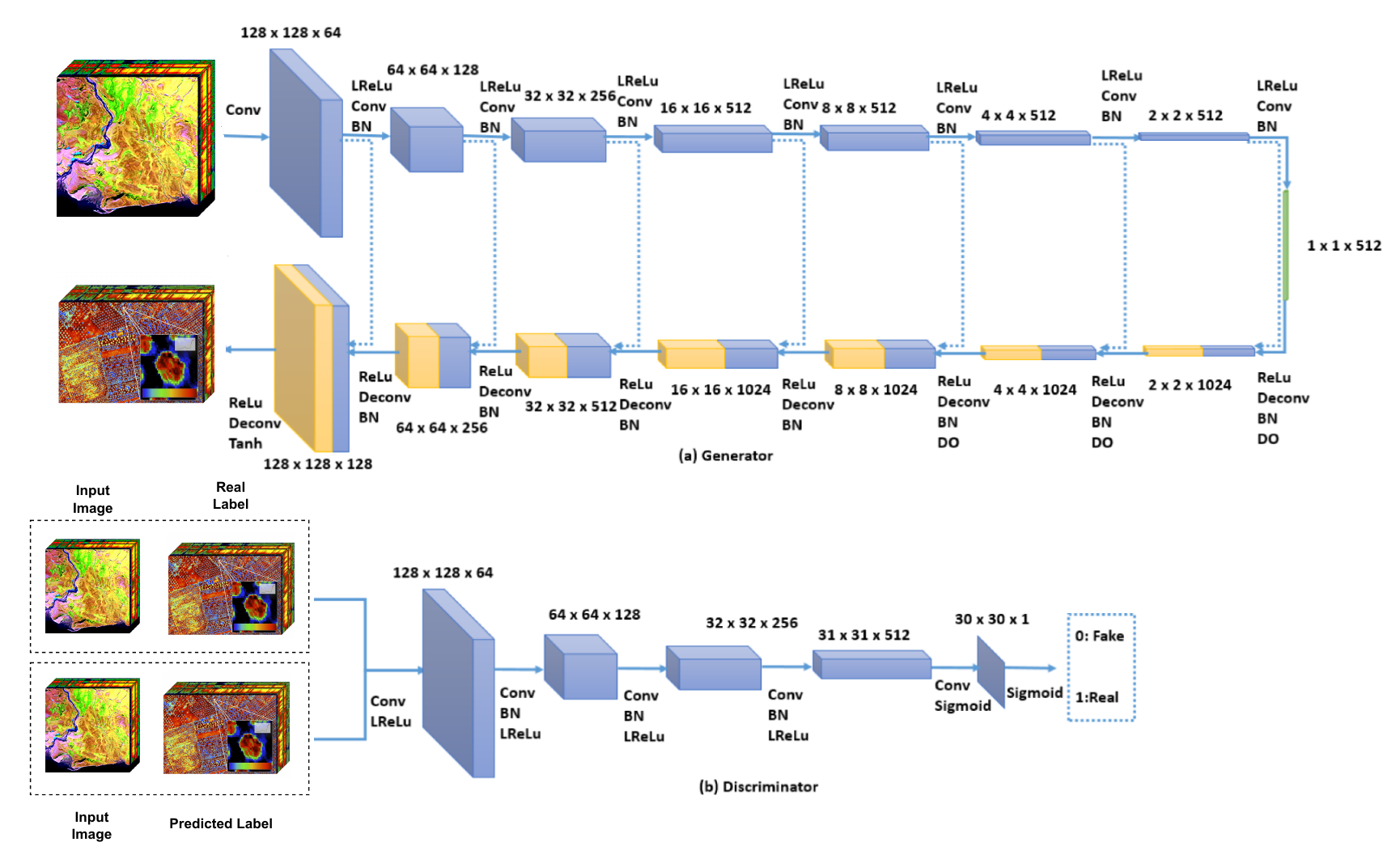}
    \caption{The architectural framework delineating the initial Generative Adversarial Network (GAN) comprises two fundamental components: (a) the generator and (b) the discriminator \cite{benjdira2020data}.}
    \label{fig:Figure10}
\end{figure}

\begin{figure}[htbp]
    \centering
    \includegraphics[width=0.93\textwidth]{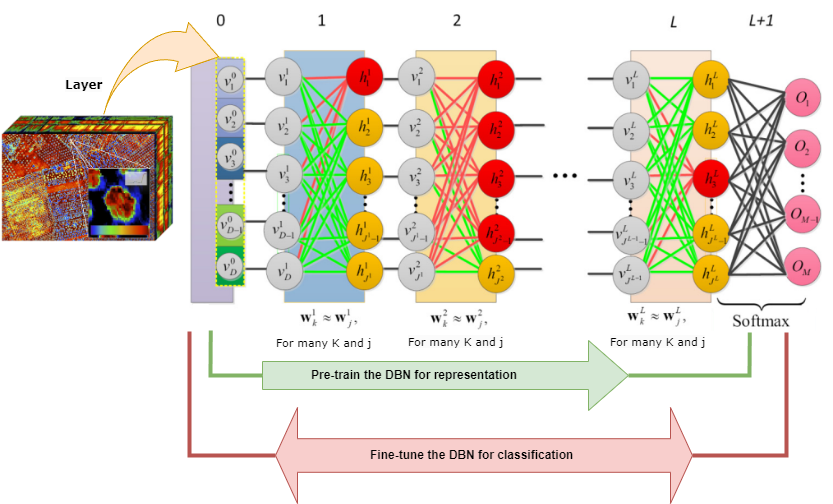}
    \caption{Graphical representation of the DBN for HSI classification \cite{zhong2017learning}.}
    \label{fig:Figure11}
\end{figure}

GANs have have been used for data augmentation, where they enable the generation of synthetic datasets that mimic real-world scenarios, thereby enhancing the robustness and accuracy of classification models. These capabilities have proven particularly valuable in agricultural applications, such as wheat disease detection \cite{goodfellow2020generative}. By reducing dependency on extensive labeled data, GANs facilitate practical and efficient solutions for agricultural analysis. Figure \ref{fig:Figure10} illustrates the proposed architecture by B. Benjdira et al. \cite{benjdira2020data}, where the generator maps target images to semantic labels, and the discriminator distinguishes real label-image pairs from generated ones during adversarial training.

Recent methods in GAN-based methodologies for HSI classification have led to the development of specialized architectures, each addressing unique challenges. Standard GANs focus on generating general synthetic hyperspectral data for augmenting datasets \cite{chen2019hyperspectral}. Conditional GANs incorporate additional constraints, such as spectral labels, to enhance control over generated data \cite{liu2020cascade}. Specialized GANs, including Capsule GAN \cite{xue2020general}, MDGAN \cite{gao2019hyperspectral}, and 3DBF-GAN \cite{he2017generative}, are optimized for handling high-dimensional spectral-spatial data in HSI. Several studies have introduced significant advancements in GAN frameworks, pushing the boundaries of their application in HSI. H. Li et al. \cite{gan1} introduced a DCGAN-based framework for enhancing unbalanced datasets in wheat crop analysis. By generating synthetic data for unsound wheat kernels, this approach improved classification accuracy from 79.17\% to 96.67\%, effectively mitigating overfitting and offering a robust solution for agricultural quality control. Ju. Wang et al. \cite{gan2} proposed the Adaptive DropBlock-Enhanced GAN, incorporating a dynamic regularization mechanism to remove feature blocks adaptively, which enhances generalization and achieves superior classification accuracy across wheat crop datasets, outperforming traditional methods. We. Zhang et al. \cite{gan3} introduced the Generative Adversarial-driven Cross-aware Network, a framework integrating a semi-supervised GAN for data augmentation with a Cross-aware Attention Network for wheat variety identification. By refining spectral, spatial, and texture features, GACNet effectively addresses data scarcity, redundancy, and noise interference. Y. Zhan et al. \cite{zhan2023semisupervised} introduced a semi-supervised GAN employing Spectral Angle Distance as a loss function, enhancing GAN convergence and generating realistic spectral samples, thus improving classification performance on benchmark datasets.

\subsection{Unsupervised Learning}
Unsupervised learning in HSI is vital for uncovering hidden patterns and structures in data without requiring labels, making it particularly suitable for large-scale datasets. This section categorizes unsupervised methods into preprocessing techniques and clustering methods. 
Preprocessing methods, such as dimensionality reduction \cite{huang2019dimensionality} and denoising \cite{pang2024hir}, enhance data quality by transforming high-dimensional hyperspectral data into more manageable and informative representations. Techniques such as DBN \cite{li2014classification, li2022manifold} and SAE \cite{windrim2019unsupervised, afrin2024enhancing} are widely used for these purposes, offering robust solutions for feature extraction, noise elimination, and spatial spectral integration.

The clustering methods aim to uncover intrinsic data structures for enabling effective classification. Advanced techniques, including diffusion models \cite{zhang2023diffucd}, unsupervised clustering algorithms such as DVIC \cite{polk2023unsupervised} and change vector analysis \cite{johnson1998change}, address challenges such as spectral mixing, segmentation, and change detection. These approaches utilize probabilistic frameworks and geometric analysis to improve spectral and spatial feature extraction, often delivering performance on par with supervised methods while significantly reducing the reliance on labeled data. Hyperspectral change detection (HCD) methods analyze temporal variations in HSI to monitor dynamic environmental or agricultural conditions, such as shifts in crop health or disease progression \cite{li2021unsupervised}. Unlike supervised approaches, unsupervised HCD frameworks \cite{liu2016unsupervised} eliminate the need for labeled training data by leveraging spectral-temporal divergence metrics, making them robust for large-scale applications like precision agriculture. For instance, these methods have been specifically applied to track wheat disease outbreaks by identifying subtle spectral anomalies linked to fungal infections  \cite{terentev2022current}, demonstrating their utility in early detection and mitigation strategies.

\subsubsection{\bfseries{Preprocessing Methods}}
DBNs, composed of multiple layers of RBMs, are highly effective for unsupervised feature learning, making them a valuable tool in HSI analysis. These networks excel at reducing the dimensionality of hyperspectral data while retaining critical spectral and spatial features \cite{li2014classification}.

Recent advancements have leveraged DBNs to enhance HSI analysis. Zh. Li et al. \cite{li2022manifold} proposed a manifold-based multi-DBN method, incorporating hierarchical initialisation to capture local geometric structures and a discrimination manifold layer to improve feature separability. This approach achieved notable classification accuracies of 78.25\%, 90.48\%, and 97.35\% on the IP, SA, and Botswana datasets, respectively, surpassing state-of-the-art methods. Efforts to advance DBNs for HSI have focused on optimizing learning efficiency and feature representation, with techniques such as manifold-based feature extraction and local global integration demonstrating their capability to handle complex hyperspectral datasets.

SAEs are widely recognized for their excellence in unsupervised feature learning, especially in HSI. These methods compress high-dimensional data into manageable representations, preserving crucial spectral and spatial features. Their adaptability to complex data structures has made SAEs instrumental in applications like dimension reduction, noise elimination, and spectral-spatial integration.

Notable advancements in SAEs for HSI include Afrin et al. \cite{afrin2024enhancing} lossy compression method that achieves high Peak Signal-to-Noise Ratios and 99\% classification accuracy on benchmark datasets. Additionally, Mou et al. \cite{mou2017unsupervised} developed a deep residual Conv–Deconv network for spectral-spatial feature extraction, demonstrating SAE's potential in unsupervised object detection. \cite{guerri2024deep} highlights that SAEs excel in reducing dimensionality while preserving critical spectral-spatial details, aiding tasks such as classification and target detection in HSI.

Significant advances in unsupervised feature learning for hyperspectral data include the work of L. Windrim et al. \cite{windrim2019unsupervised}, who proposed SID-SAE and SA-SAE—specialized autoencoders designed for dimensionality reduction. These methods integrate remote-sensing spectral similarity measures into their loss functions, with SID-SAE employing Spectral Information Divergence to assess probabilistic similarity and SA-SAE using Spectral Angle to emphasize shape over magnitude. By preserving critical spectral properties, these methods improve class separability and support clustering and classification tasks, particularly in scenarios with limited labeled data. To evaluate their effectiveness, the low-dimensional features were clustered using k-means, with the number of clusters corresponding to the dataset classes, and clustering performance measured using the Adjusted Rand Index. Additionally, the unsupervised features were assessed with supervised classifiers such as Support Vector Machines, k-Nearest Neighbors, and Decision Trees, highlighting the robustness of these methods across diverse evaluation strategies.

L. Pang et al. \cite{pang2024hir} proposed HIR-Diff for unsupervised HSI restoration. Using pre-trained diffusion models with data fidelity terms, total variation priors, and low-rank approximation (SVD and RRQR), HIR-Diff addresses tasks like denoising and super-resolution, achieving efficient and accurate restoration. 
The model has a robust generative approach, is data-efficient (no paired training data), is adaptable across tasks with high computational costs, and has dataset-specific parameter tuning.

\subsubsection{\bfseries{Clustering Methods (Classification)}}

Clustering methods for HSI focus on grouping pixels based on spectral-spatial similarities, enabling effective classification without labeled data. Recent advancements have integrated DL techniques such as diffusion models and SAEs into unsupervised segmentation frameworks, enhancing feature extraction and clustering accuracy. Additionally, anomaly change detection methods have emerged as a key approach for identifying temporal changes in hyperspectral data, further advancing the capabilities of unsupervised learning in dynamic environments. These techniques improve the robustness and scalability of clustering methods, facilitating more efficient analysis of high-dimensional HSI datasets.

Notable advancements in diffusion models for HSI include X. Zhang et al. \cite{zhang2023diffucd} introducing the DiffUCD framework for hyperspectral change detection. This approach employs a Semantic Correlation Diffusion Model to extract spectral-spatial features and a Cross-Temporal Contrastive Learning mechanism to align spectral representations across time. SCDM enhances meaningful feature representation while addressing environmental changes and imaging inconsistencies. Meanwhile, CTCL mitigates pseudo-changes and improves feature invariance by leveraging abundant unlabeled samples, eliminating reliance on labeled data, and reducing the need for labor-intensive annotation. DiffUCD demonstrates the potential of diffusion models in remote sensing, achieving competitive results against supervised techniques while enhancing labeling efficiency.

S. L. Polk et al. \cite{polk2023unsupervised} developed D-VIC, an unsupervised clustering algorithm for HSI that integrates high-dimensional geometry and spectral abundance. Using diffusion distances and pixel purity, D-VIC identifies cluster modes representative of material structures in spectrally mixed HSIs. A modular spectral unmixing step combines HySime, AVMAX, and a non-negative least-squares solver to effectively estimate material abundances and purities. Experimental results reveal that D-VIC outperforms state-of-the-art methods on benchmark datasets, demonstrating robustness to hyperparameter tuning and scalability to large HSIs. However, the method remains computationally intensive because of spectral unmixing and diffusion-difference computations. 

A. Mughees and L. Tao \cite{mughees2016efficient} introduced a method that integrates deep learning with unsupervised segmentation. SAEs were used to extract spectral features, which were then combined with spatial information using a boundary adjustment technique and a maximum voting approach, achieving significant improvements in spectral-spatial classification accuracy on standard hyperspectral datasets.

In hyperspectral anomaly change detection, several methods aim to improve accuracy and reduce reliance on labeled data by tackling challenges in feature extraction, anomaly representation, and computational efficiency. For instance, BCG-Net \cite{hu2023binary}, an unsupervised multi-task framework, enhances detection accuracy in both binary and multiclass tasks. Similarly, DSFA \cite{du2019unsupervised} integrates DL with slow feature analysis to emphasize changes, surpassing other models in binary detection performance.

Building on these methods, M. Hu. \cite{hu2022hypernet} introduces HyperNet, a self-supervised framework designed for hyperspectral change detection. HyperNet features a spatial-spectral attention module for precise pixel-level extraction and a focal cosine loss function that optimizes training by balancing easy and hard samples. Unlike patch-based models, it processes entire images, improving generalization across datasets. However, HyperNet faces challenges in computational complexity and sensitivity to noisy data, which require further exploration.


\subsection{Discussion}
DL models have significantly advanced the field of HSI analysis, offering powerful tools for classification, segmentation, and interpretation across diverse datasets. These models, while diverse in architecture and function, share the common goal of enhancing data analysis and accuracy, particularly in applications such as wheat crop health monitoring and disease detection. Their performance, as summarized in Table \ref{tab:Table 6} and Table \ref{tab:Table 7}, highlights both their strengths and limitations in various HSI tasks.

CNNs stand out as the most widely used DL models for HSI analysis, with their ability to automatically extract intricate spectral and spatial features from hyperspectral data cubes. CNNs demonstrate exceptional classification accuracy, achieving up to 99.9\% on datasets like PU and SA. Their strength lies in the layered structure, which efficiently captures both spatial and spectral hierarchies, making CNNs ideal for tasks such as land use classification and disease detection in wheat crops. However, while CNNs excel in general classification tasks, their reliance on large labeled datasets can be a limitation in scenarios where annotated data are scarce.

Moving beyond static spatial-spectral analysis, RNNs offer a unique advantage in handling sequential data by capturing temporal dependencies between spectral bands. This capability is particularly useful in agricultural applications where spectral data vary over time, such as in the tracking of wheat crop growth or monitoring environmental stressors. RNNs perform well with accuracies between 88.6\% and 97.5\%, but they often struggle with computational efficiency and scalability when processing long sequences, especially in high-dimensional hyperspectral data.

In contrast, GANs address a critical challenge in HSI analysis: the limited availability of labeled data. By generating synthetic samples, GANs improve data augmentation and enable more robust training for classification models. However, the performance of GANs can vary widely, with accuracies ranging from 75.6\% to 99.8\%. While effective in data augmentation and enhancing model robustness, GANs are sensitive to training instability and often require careful tuning to prevent overfitting, particularly in complex agricultural datasets like those involving wheat crops.

DBNs and SAEs both excel in unsupervised feature learning, offering powerful methods for dimensionality reduction while retaining essential spectral and spatial features. DBNs, with accuracies ranging from 78.2\% to 98.9\%, are particularly useful in scenarios with limited labeled data, allowing for effective feature extraction without extensive supervision. SAEs also perform well in tasks such as the detection of DON levels in wheat kernels, achieving up to 98.5\% accuracy. These models are instrumental in automated feature extraction, but their reliance on multi-layered architectures can lead to longer training times and higher computational costs. Diffusion models have also emerged as promising tools for HSI analysis, particularly for image restoration and anomaly detection tasks. By simulating noise addition and removal processes, diffusion models effectively capture spectral-spatial dependencies in hyperspectral data, enhancing the quality of low-resolution or noisy images. 

Among the state-of-the-art models, transformer architectures have emerged as particularly effective in HSI classification because of their ability to model extensive spectral relationships across bands. Achieving up to 99.8\% accuracy, transformer models are particularly well suited for complex tasks like wheat crop classification, where subtle variations across nonadjacent bands can be critical. The Mamba module further enhances the robustness of HSI analysis by integrating spatial and spectral information more effectively. With accuracy rates exceeding 99\% in some cases, Mamba models are highly effective for wheat crop monitoring and disease detection, offering a computationally efficient alternative to traditional models.

Future advancements in DL models for HSI analysis should prioritize computationally efficient and lightweight architectures capable of effectively scaling to large datasets, as highlighted for CNNs, RNNs, and SAEs. The integration of attention mechanisms in RNNs and Transformer-based architectures could significantly enhance the capture of long-range spectral-spatial dependencies, improving real-time processing and generalization across diverse agricultural applications. Hybrid approaches, such as combining DBNs with other DL paradigms and innovations such as advanced loss functions and domain-specific attention modules in Transformers, may further optimize performance and robustness. Additionally, the exploration of SAE and DBN models with applications in large-scale agricultural monitoring, real-time analysis, and improved spectral-spatial integration represents a promising avenue for innovation in HSI analysis. 

In summary, while CNNs and Transformers provide the highest accuracy and robustness in HSI tasks, other models like RNNs, GANs, DBNs, and SAEs each contribute uniquely depending on the application. CNNs and Transformers dominate in classification and spatial-spectral analysis, whereas RNNs and GANs are better suited for sequential data processing and augmentation. Using the strengths of each model, researchers can address the unique challenges of HSI analysis in agriculture, particularly for the monitoring of wheat crop health and the prediction of yields. The selection of the appropriate model depends on the specific requirements of the task, such as data availability, computational efficiency, and the complexity of the spectral-spatial relationships in the dataset.

\begin{table}[htbp]
\caption{A summary of the results obtained from DL methods, including CNN, RNN, GAN, and DBN, applied to HSI. The meaning of abbreviations used in this table is: OA: Overall Accuracy, IP: Indian Pines, Accuracy= acc}

\begin{tabular}{lllp{4cm}p{5cm}}
\cline{2-5}
\hline
Learning paradigm                         & Method                       & Article                                             & Application                                                                                             & Results                                                                                           \\ \cline{2-5} \hline
\multirow{35}{*}{Supervised} & \multirow{13}{*}{CNN}        & \cite{simon2022convolutional}      & Land use classification (LUC)                                                                           & OA=97\%                                                                                           \\ \cline{3-5} 
                                      &                              & \cite{sadeghi2021neural}           & Classification                                                                                          & OA=98.6\%                                                                                         \\ \cline{3-5} 
                                      &                              & \cite{dhakal2023machine}           & Fusarium Head Blight (FHB) disease classification                                                       & OA=97\%                                                                                           \\ \cline{3-5} 
                                      &                              & \cite{chen2016deep}                & Classification                                                                                          & OA:IP=87.8\%, wheat crop class acc= 98.64\%                                                                \\ \cline{3-5} 
                                      &                              & \cite{li2017spectral}              & Classification                                                                                          & OA:IP=87.8\%, wheat crop class acc= 99.82\%                                                                 \\ \cline{3-5} 
                                      &                              & \cite{yang2018hyperspectral}       & Classification                                                                                          & \begin{tabular}[c]{@{}l@{}}OA:IP=99.4\%, wheat crop class acc= 100\%  \end{tabular}           \\ \cline{3-5} 
                                      
                                      &                              & \cite{song2018hyperspectral}       & Classification                                                                                          & OA:IP=98.5\%, wheat crop class acc= 99.50\%                                                                  \\ \cline{3-5} 
                                      &                              & \cite{gong2019novel}               & Classification                                                                                          & OA:IP=98.9\%, wheat crop class acc= 98\%                                                                          \\ \cline{3-5} 
                                      &                              & \cite{bing2019deep}                & Classification                                                                                          & OA:IP=93.1\%                                                              \\ \cline{3-5} 
                                      &                              & \cite{lee2017going}                & Classification                                                                                          & OA:IP=93.6\%                                                                 \\ \cline{3-5} 
                                      &                              & \cite{roy2019hybridsn}             & Classification                                                                                          & AO:IP=98.3\%                                                                 \\ \cline{2-5} 
                                     
                                      & \multirow{5}{*}{RNN}         & \cite{mou2017deep}                 & Classification                                                                                          & OA:IP=88.6\%, wheat crop class acc= 59.56\%                                                             \\ \cline{3-5} 
                                      &                              & \cite{rnn1}                        & Classification                                                                                          & OA: 97.5\%                                                                                        \\ \cline{3-5} 
                                      &                              & \cite{rnn3}                        & Classification                                                                                          & OA: 89.6\%                                                                                        \\ \cline{3-5} 
                                      &                              & \cite{venkatesan2019hyperspectral} & Classification                                                                                          & Recall= 88\%                                                                                      \\ \cline{3-5} 
                                      &                              & \cite{luo2018shorten}              & Classification                                                                                          & OA:IP=90.3\%                                                                            \\ \cline{2-5} 
                                      & \multirow{4}{*}{Transformer}  & \cite{tran1}                       & Classification                                                                                          & OA:IP=99.06\%                                                            \\ \cline{3-5} 
                                      &                              & \cite{tran2}                       & Classification                                                                                          & OA:IP=96.8\%, wheat crop class accuracy= 98.64                                                                \\ \cline{3-5} 
                                      &                              & \cite{tran3}                       & Classification                                                                                          & OA:IP=98.4\%, wheat crop class accuracy= 100\%                                                                                      \\ \cline{2-5} 
                                      & \multirow{5}{*}{Mamba}       & \cite{he20243dss}                  & Classification                                                                                          & \begin{tabular}[c]{@{}l@{}}OA: IP=96.47\%\end{tabular}                   \\ \cline{3-5} 
                                      &                              & \cite{sheng2024dualmamba}          & Classification                                                                                          & \begin{tabular}[c]{@{}l@{}}OA: IP=99.23\%, wheat crop class acc= 99.73\%\end{tabular} \\ \cline{3-5} 
                                      &                              & \cite{yang2024graphmamba}          & Classification                                                                                          & \begin{tabular}[c]{@{}l@{}}OA: IP=96.43\%\end{tabular}                  \\ \cline{3-5} 
                                      &                              & \cite{yang2024hsimamba}            & Classification                                                                                          & \begin{tabular}[c]{@{}l@{}}OA: IP=89.92\%, wheat crop class acc= 99.65\%\end{tabular}                \\ \cline{3-5} 
                                      &                              & \cite{zhou2024mamba}               & Classification                                                                                          & \begin{tabular}[c]{@{}l@{}}OA: IP=92.07\%\end{tabular} \\ \cline{2-5} 
                                      
                                      & \multirow{2}{*}{SAE}         & \cite{sae1}                        & Detection of DON levels in FHB wheat kernels and wheat flour & OA: Vis-NIR=98.5\% / 91.5\%, SWIR= 95.7\% / 97.1\%                                                \\ \cline{3-5}  
                                      &                              & \cite{xing2016stacked}             & Classification                                                                                          & OA:IP=92.06\%                                                              \\ \cline{2-5} 
                                      & \multirow{3}{*}{DBN}           & \cite{dbn1}                        & Classification                                                                                          & OA:IP=96.2.6\%, wheat crop class acc= 91.67\%                                                                         \\ \cline{3-5} 
                                      &                              & \cite{sellami2019spectra}          & Classification                                                                                          & OA:IP=94.9\%, wheat crop class acc= 93.91\%                                                                           \\ \cline{3-5} 
                                      &                              & \cite{yang2019feature}             & Classification                                                                                          & OA:IP=97.9\%                                                 \\ \cline{2-5} \hline

\end{tabular}
\label{tab:Table 6}
\end{table}


\begin{table}[htbp]
\caption{A summary of the results obtained from DL methods, including SAE and Transformer, applied to HSI. The meaning of abbreviations used in this table are: OA: Overall Accuracy, IP: Indian Pines, SB: Santa Barbara, BA: Bay Area, He: Hermiston, JR: Jasper Ridge, SWIR (Short-Wave Infrared), and VNIR (Very Near Infrared).}

\begin{tabular}{p{2.5cm}lllp{5cm}}
\cline{2-5}
\hline
Learning paradigm                  & Method                     & Article                                         & Application                                                                                             & Results                                                                                      \\ \cline{2-5} \hline
\multirow{14}{*}{Semi-Supervised} & \multirow{5}{*}{Pseudo-Labeling} & \cite{zhao2024enhancing}                        & Classification                                                                                          & OA:IP=97.41\%, wheat crop class acc= 100\%                                                                    \\ \cline{3-5} 
                               &                            & \cite{wang2021improved}                         & Classification                                                                                          & OA:IP=98.20\%, wheat crop class acc= 98.91\%                                                          \\ \cline{3-5} 
                               
                               &                            & \cite{fang2018semi}                             & Classification                                                                                          & OA:IP=97.66\%, wheat crop class acc= 96.04\%                                                                \\ \cline{3-5} 
                               &                            & \cite{zhang2020semi}                            & Classification                                                                                          & OA:IP=95.14\%   \\ \cline{2-5}
& \multirow{9}{*}{GAN}       & \cite{zhan2017semisupervised}                  & Classification                                                                                          & OA:IP=83.5\%                                                                                 \\ \cline{3-5} 
                               &                            & \cite{zhu2018generative}                        & Classification                                                                                          & OA:IP=89\%, wheat crop class acc= 49.75\%                                                            \\ \cline{3-5} 
                               &                            & \cite{xue2020general}                           & Classification                                                                                          & OA:IP=99.1\%, wheat crop class acc= 98.69\%                                                                        \\ \cline{3-5} 
                              
                               &                            & \cite{gao2019hyperspectral}                     & Classification                                                                                          & OA:IP=95.7\%, wheat crop class acc= 100\%                                                              \\ \cline{3-5} 
                               &                            & \cite{he2017generative}                         & Classification                                                                                          & OA:IP=75.6\%, wheat crop class acc= 99.41\%                                                            \\ \cline{3-5} 
                               &                            & \cite{gan1}                                    & Classification of wheat kernels                                                                         & OA:IP=96.6\%                                                                                  \\ \cline{3-5} 
                               &                            & \cite{gan2}                                    & Classification                                                                                          & OA:IP=97.2\%, wheat crop class acc= 97.08\%                                                             \\ \cline{3-5} 
                               &                            & \cite{benjdira2020data}                         & Segmentation                                                                                            & OA1: 58.8\%, OA2: 36.3\%                                                                     \\ \cline{2-5} \hline
                                                                                             
\multirow{5}{*}{Unsupervised}                                                       
                                                              & \multirow{2}{*}{SAE}       & \cite{windrim2019unsupervised}                  & Classification                                                                                          & Clustering Result= 70\%                                                                      \\ \cline{3-5} 
                               &                            & \cite{mughees2016efficient}                     & Classification                                                                                          & Acc:IP=98.6\%                                                                     \\ \cline{2-5}
 
                               & \multirow{2}{*}{Diffusion} & \cite{zhang2023diffucd}                         & Classification                                                                                          & \begin{tabular}[c]{@{}l@{}}OA:SB=96.87\%,BA=96.35\%,\\ He=95.47\%\end{tabular}               \\ \cline{3-5} 
                               &                            & \cite{polk2023unsupervised}                     & Classification                                                                                          & OA:IP=44.5\%                                                            \\ \hline
\end{tabular}
\label{tab:Table 7}
\end{table}

\section{Applications of HSI in wheat crops}
\label{sec:Section 6}

\begin{figure}[htbp]
    \centering    \includegraphics[width=0.82\linewidth]{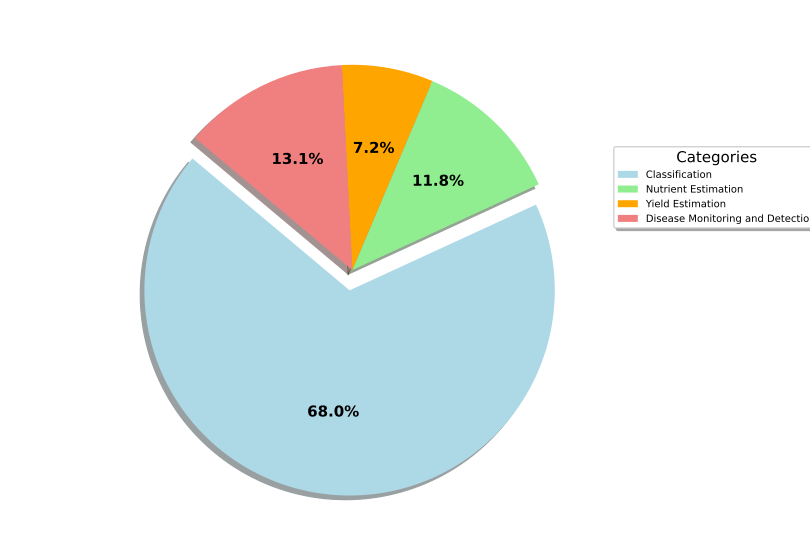}
    \caption{Number of Published Articles by Year on DL with Hyperspectral Data in Wheat Crops.}.
    \label{fig:Figure13}
\end{figure}

Extensive research on HSI has demonstrated the high reliability of DL algorithms, particularly in agricultural applications for wheat crops. The pie chart in Figure \ref{fig:Figure13} illustrates the distribution of research efforts across four key categories. Classification dominates at 68.0\%, underscoring its fundamental role as the backbone of HSI-based wheat analysis. Disease Detection and Monitoring follows at 13.1\%, reflecting its critical yet complementary function in ensuring crop health. Nutrient Estimation, at 11.8\%, signifies a notable focus on optimizing wheat growth and productivity. Finally, Yield Estimation, accounting for 7.2\%, represents a more specialized but still relevant area of research.
This distribution highlights classification as the primary driver of research, while disease monitoring and nutrient estimation serve as essential supporting components. Despite receiving comparatively less attention, yield estimation remains integral to a comprehensive approach to wheat cultivation. The following applications will be considered in this section: Wheat Crop Classification, Wheat Crop Disease Detection and Monitoring, Wheat Crop Nutrient Estimation, and Wheat Crop Yield Estimation.

\subsection{Wheat Crop Classification}
Agricultural crop classification has made substantial use of HSI, which enables precise crop discrimination through detailed spectral information. Classifying wheat harvests and differentiating between wheat varieties, grain types, and physiological features are among the applications where HSI has been instrumental \cite{madhavan2023wheat}. By analyzing subtle spectral variations, HSI allows for identification of chlorophyll content, leaf structure, and other physiological traits specific to each wheat variety \cite{sethy2022identification,site4}. These advancements have led to the development of robust DL models that significantly improve classification accuracy.

Table \ref{tab:Table 8} summarizes key research efforts on HSI-based wheat classification, showcasing the diverse range of deep learning techniques applied in this domain. Studies have demonstrated the effectiveness of CNNs, multi-layer perceptrons (MLPs), and hybrid DL approaches in achieving high classification accuracy. As research continues to evolve, integrating large-scale datasets and optimizing model architectures remains crucial for improving scalability and performance.

E. Dreier et al. \cite{dreier2022hyperspectral} demonstrated the effectiveness of deep CNNs for wheat grain classification, showing that these models outperform traditional approaches in recognizing wheat grain variations. Similarly, P. Sethy et al. \cite{sethy2022identification} employed a Neural Network Pattern Recognizer to classify wheat grains using a dataset from the University of California, Irvine’s machine learning repository, achieving a classification accuracy of 96.7

In addition to spectral-based classification, geometric characteristics have also been utilized to enhance classification performance. S. Lingwal et al. \cite{lingwal2021image} demonstrated that the shape and structural characteristics of wheat grains could be effectively leveraged to improve classification accuracy. Expanding on this, K. Sabanci et al. \cite{sabanci2017computer} developed an MLP-based computer vision system to classify wheat grains into bread and durum varieties, integrating high-resolution imaging to capture spectral and textural differences, thereby improving the reliability of wheat-type identification.

Further advances in DL-based hyperspectral classification were introduced by W. Hu et al. \cite{hu2015deep}, who designed a five-layer deep CNN for spectral data analysis. Their model demonstrated superior performance compared to traditional classifiers such as support vector machines (SVMs), underscoring the advantage of deep learning in hyperspectral wheat classification. V. Slavkovikj et al. \cite{slavkovikj2015hyperspectral} proposed a feature learning approach using CNNs, where their model automatically extracted band-pass features from hyperspectral data. Applied to a widely used remote sensing hyperspectral dataset, their method showed strong classification performance across multiple crops, including wheat.

The integration of large-scale datasets has further propelled wheat classification research. E. David et al. \cite{david2020global} utilized the Global Wheat Head Dataset (GWHD) to analyze wheat attributes, employing AlexNet with multi-feature fusion to improve classification accuracy. Their work contributed to wheat breeding research and agricultural classification, demonstrating the applicability of deep learning to real-world large datasets.

Beyond hyperspectral imaging, Near-Infrared Spectroscopy (NIR) has also been explored for wheat classification. N. Caporaso et al. \cite{caporaso2018near} investigated the role of HSI and NIR in cereal grain assessment, showing that these technologies, when combined with deep learning models such as CNNs and MLPs, can achieve high classification accuracy. Their study highlighted the benefits of non-destructive, real-time quality assessment, which aids in fungal infection detection and grain composition analysis.

While significant progress has been made, challenges such as high computational demands, the need for large labeled datasets, and spectral variability remain areas for further research. Future advancements in model optimization and dataset augmentation techniques could enhance the scalability and accuracy of HSI-based wheat classification models.

\begin{table}[htbp]
\caption{An overview of the key research conducted on wheat crop classification using HSI.}
\begin{tabular}{lp{5cm}lp{3.5cm}l}
\hline
Year & Objective & Method & Results \\
\hline
2019 \cite{bao2019rapid} & Classification of wheat variety using HSI and chemometrics & PCA, LDA & Accuracy=91.3\% \\
\hline
2021\cite{ozkan2021wheat} & DL system for wheat samples classification & AlexNet FC6, VGG16 FC7 & AlexNet FC6: 96.00\%, VGG16 FC7: 99.00\% \\
\hline
2022 \cite{yipeng2022determination} & Classification of wheat kernels damaged by FHB using EWs from HSI images & ASSDN & SetT: 100\%, SetP: 98.31\%, AUC: 99.85\% \\
\hline
2023 \cite{zhu2023identification} & Classification model for online detection of sound and sprouted wheat kernels & 1D CNN, 2D CNN, 3D CNN & 2D CNN= 96.02\%, 3D CNN= 98.4\% \\
\hline
2023 \cite{que2023identification} & Non-destructive method for wheat seed variety identification & CNN & Accuracy=100\% \\
\hline
\end{tabular}
\label{tab:Table 8}
\end{table}

\subsection{Wheat Crop Disease Detection and Monitoring}
The occurrence of diseases in wheat crops presents significant challenges for farmers, as untreated infections can lead to severe yield losses. Fast and non-destructive, HSI has demonstrated notable results in diagnosing plant diseases by capturing high-resolution spectral data across visible to near-infrared wavelengths \cite{wan2022hyperspectral}. This technology provides comprehensive insights into the physiological and metabolic changes in plants, enabling early disease detection before symptoms become visible \cite{terentev2023hyperspectral}.

Table \ref{tab:Table 11} summarizes recent advancements in the detection of wheat disease based on DL, highlighting the potential of HSI in precision agriculture. These studies demonstrate how HSI technology, combined with DL models, enables early disease detection, classification, and monitoring, ultimately improving wheat crop management.

G. Feng et al. \cite{feng2024wheat} showed that HSI, by analyzing the interaction between light and plant tissues, can capture subtle spectral differences, effectively identifying biotic and abiotic stresses. This capability makes HSI a powerful tool for detecting wheat diseases in their early stages. Similarly, X. Jin et al. \cite{jin2018classifying} used deep neural networks to classify HSI pixels, improving disease detection accuracy in wheat fields.

The integration of DL algorithms further strengthens the diagnostic capabilities of HSI by training models to distinguish healthy and diseased wheat crops according to spectral signatures \cite{terentev2023hyperspectral}. W. Khotimah et al. \cite{khotimah2023mce} evaluated the MCE-ST network, a DL-based model applied to cassava disease and wheat salt stress datasets, demonstrating its effectiveness in objectively assessing disease severity.

By combining HSI technology with DL models, researchers can develop robust disease monitoring frameworks that allow for timely intervention and optimized disease management strategies. This fusion of technologies enhances agricultural productivity and contributes to global food security.
\begin{table}[htbp]
    \caption{A summary of the important studies on HSI's use for wheat crop disease monitoring and detection.}
    \small
    \begin{tabular}{p{2.5cm}p{4cm}p{4cm}p{3cm}}
    \hline
    Year & Objective & Method & Results \\
    \hline
    2021\cite{khan2021early}  & Early detection of powdery mildew disease in wheat using HSI & Partial least-squares regression (PLSR) & Accuracy= 82.35\%, (R2 =0.748-0.722) \\
    \hline
    2022 \cite{almoujahed2022detection} & Detection of fusarium head blight (FHB) in wheat using a hyperspectral camera & Support Vector Machine (SVM), Artificial Neural Network (ANN), Logistic Regression (LR) & Accuracy= 95.6\% \\
    \hline
    2022 \cite{rangarajan2022detection} & DL models for the detection of fusarium head blight (FHB) in wheat & CNN model, DarkNet 19 & F1 score =100\% \\
    \hline
    2022 \cite{feng2022hyperspectral} & Hyperspectral monitoring of powdery mildew disease severity, improved monitoring accuracy of wheat powdery mildew & CARS algorithm, SPA algorithm & R2 =0.733-0.836 \\
    \hline
    2022 \cite{yipeng2022determination} & HSI and DL for FHB-damaged wheat kernel classification. Architecture self-search deep network achieved high accuracy & KNN (K-Nearest Neighbors), SVM (Support Vector Machine), CNN, ASSDN (Architecture Self-Search Deep Network) & AUC= 0.9985 \\
    \hline
    2023 \cite{xu2023wheat} & Proposed DL algorithm for wheat leaf disease identification. Outperforms existing CNN models in recognition precision. & Lightweight Detection & Accuracy =99.95\% \\
    \hline
    2023 \cite{sereda2023development} & Develop remote methods for diagnosing the state of wheat crops and monitoring diseases using spectral equipment and original monitoring tools. & satellite and UAV & Reflectance Values, Correlation with Disease, Effectiveness of Remote Sensing \\
    \hline
    2023 \cite{blasch2023potential} & The paper evaluates the use of very high-resolution satellite (VHRS) and UAV imagery for the early detection and phenotyping of yellow rust (YR) and stem rust (SR) in wheat crops in Ethiopia. It focuses on how different wheat varieties respond to disease stress. & VHRS imagery, NDVI, RVI, Disease Index (DI).& Accuracy =85.2\% \\
    \hline
    2024 \cite{feng2024wheat} & Provide a comprehensive review of advanced automatic non-destructive detection techniques for Fusarium head blight (FHB) in wheat. & Lightweight Detection model  & Accuracy =94.3\% \\
    \hline
    2024 \cite{chang2024recognition} & The objective of the research is to achieve higher classification accuracy in detecting wheat rust diseases using DL models. & DenseNet121 and Imp-DenseNet. & Accuracy =98.32\% \\
    \hline
    \end{tabular}
    \label{tab:Table 11}
\end{table}

\subsection{Wheat Crop Nutrient Estimation}
Assessing nutrient levels in wheat crops is crucial for optimizing agricultural practices and ensuring healthy growth. Nutritional deficiencies can significantly impact crop yield and quality, posing challenges for farmers. HSI technology offers a promising approach for monitoring the nutritional status of wheat by analyzing distinct spectral signatures that provide insights into the quality and health of wheat grains. Recent advancements, particularly in DL-based methods, have further enhanced the accuracy of these assessments. Table \ref{tab:Table 9} provides an overview of key studies that have explored DL-based nutritional estimation of wheat crops, showcasing the integration of HSI and computational models for improved nutrient analysis.

Recent research \cite{ma2022applications} has demonstrated the effectiveness of HSI in assessing wheat grain quality. By capturing and analyzing spectral data, critical details about the nutritional makeup and general health of wheat kernels can be obtained. The integration of HSI with DL algorithms has been extensively investigated, particularly concerning data gathered by drones \cite{barbedo2023review}. This combination allows for advanced analysis of HSI, enabling more precise estimations and forecasts regarding crop health.

N. Hu et al. \cite{hu2021predicting} proposed a quick and cost-effective method for determining the micronutrient content of wheat grains using HSI. They posited that HSI could serve as a useful tool for predicting micronutrient levels in wheat, leading to timely interventions for nutrient management. This research utilized a portable Near-Infrared (NIR) device for mixing powdered food samples and focused on estimating micronutrient content effectively.

Another paper by Y. Song et al. \cite{song2021assessment} aimed to identify the best color characteristics and models for detecting chlorophyll concentration in wheat. The research involved multiple steps, including target extraction, feature extraction, feature selection, and modeling. The proposed method for assessing chlorophyll content showed high agreement with conventional chlorophyll meters during trials in a wheat experimental field, demonstrating its reliability and effectiveness compared to other regression models.

In general, the use of digital image analysis in agricultural practices highlights the importance of evaluating crop nutrition. By leveraging HSI technology, farmers can make informed decisions regarding fertilization and nutrient management, ultimately leading to improved crop yields and sustainability.
\begin{table}[htbp]
\caption{A summary of the important studies on HSI's use in wheat crop nutrient estimation. Abbreviations: ML (Linear Model), OPM (2nd Order Polynomial Mode), EM (Exponential Model), LAI (leaf area index), and LCC (leaf chlorophyll content).}
\begin{tabular}{p{2.5cm}p{4cm}p{4cm}p{3cm}}
\hline
Year & Objective & Method & Results \\
\hline
2021 \cite{song2021assessment}  & Feature extraction and model building for crop nutrition measurement. Proposed SBRR approach for estimating wheat chlorophyll content & Stepwise-based ridge regression (SBRR) & Optimal SBRR model: R2 = 0.718, RMSE = 5.111; Verification results: R2 = 0.794, RMSE = 4.304 \\
\hline
2021 \cite{yang2021estimation} & Estimation of leaf nitrogen content in wheat & PLS, GBDT, SVR & R2 = 0.975 for calibration set, R2 = 0.861 for validation set \\
\hline
2021 \cite{astaoui2021mapping} & Monitoring wheat crop using UAV multispectral imagery. Estimating wheat yield and biophysical parameters. Mapping wheat dry biomass and nitrogen uptake & Random Forest technique and linear regression model & LM: R2 = 0.761, RMSE = 0.63; OPM: R2 = 0.741, RMSE = 0.67; EM: R2 = 0.690, RMSE = 0.67 \\
\hline
2022 \cite{lu2022assessment} & UAV-based multi-view images improve the estimation of nitrogen nutrition & Support Vector Regression (SVR), Extreme Learning Machine (ELM), Random Forest (RF) & LNC: R2 = 0.61, RMSE = 0.37\%; PNC: R2 = 0.52, RMSE = 0.24\%; LNA: R2 = 0.72, RMSE = 1.45 g/m\textsuperscript{2}; PNA: R2 = 0.62, RMSE = 3.02 g/m\textsuperscript{2} \\
\hline
2023 \cite{du2023incremental} & Develop a universal method for accurate estimation and diagnosis of leaf chlorophyll content (LCC) and leaf area index (LAI), and diagnose nitrogen levels in crops & Bi-objective constrained deep neural network (DNNCA model) & LAI: R2 = 0.64-0.82, RMSE = 0.58-1.02 m\textsuperscript{2}/m\textsuperscript{2}; LCC: R2 = 0.56-0.82, RMSE = 3.9-10.5 ug/cm\textsuperscript{2} \\
\hline
\end{tabular}
\label{tab:Table 9}
\end{table}

\subsection{Wheat Crop Yield Estimation}
Accurately estimating wheat crop yield is essential for effective agricultural planning and management. Reliable yield predictions enable farmers to allocate resources efficiently, optimize their practices, and improve overall productivity. Recent advancements in HSI and DL have significantly enhanced yield estimation accuracy. By integrating hyperspectral data with other remote sensing sources, researchers have developed more precise predictive models. Table \ref{tab:Table 10} summarizes key studies focusing on DL-based wheat crop yield estimation, illustrating various methodologies and technological advancements in this area.

To predict agricultural yield at a spatial resolution of 5 meters by 5 meters, E. Cheng et al. \cite{cheng2022wheat} combined gridded yield data with DL algorithms, utilizing Sentinel-2 multispectral and ZY-1 02D hyperspectral data. This integration enabled precise yield predictions, offering valuable insights into expected crop output. Similarly, A. Moghimi et al. \cite{moghimi2020aerial} emphasized that UAVs equipped with hyperspectral cameras have facilitated rapid and cost-effective yield estimation. This UAV-based approach allows high-resolution aerial imagery to be captured, significantly benefiting winter wheat breeding programs \cite{liu2023estimation}.

 researchers have employed low-altitude UAV-based hyperspectral data collection at various points in the winter wheat crop canopy to refine yield predictions \cite{li2022uav}. Additionally, the integration of open-source computational tools such as R and Python, combined with automated hyperspectral narrowband vegetation index analysis, has demonstrated promising results in crop yield estimation \cite{li2022toward}. This method harnesses the power of DL algorithms to process HSI data, extracting meaningful insights for yield forecasting.

By leveraging HSI and DL-driven techniques, farmers can obtain more accurate and timely yield predictions, leading to improved decision-making, better resource management, and higher crop productivity.

\begin{table}[htbp]
    \small
    \caption{A summary of the important studies on HSI's use for crop yield estimate of wheat crops.}
    \begin{tabular}{p{2.5cm}p{4cm}p{4cm}p{3cm}}
    \hline
    Year & Objective & Method & Results \\
    \hline
    2020 \cite{moghimi2020aerial} & Automated framework for selecting advanced crop varieties. Hyperspectral camera mounted on unmanned aerial vehicle & Deep neural networks (DNN model). & Predicting yield: 0.79 (sub-plot scale), 0.41 (plot scale) \\
    \hline
    2022 \cite{cheng2022wheat} & Used machine learning to estimate wheat yields & LSTM (Long Short-Term Memory), RF (Random Forest), GBDT (Gradient Boosting Decision Tree), SVR (Support Vector Regression). & LSTM model (R2=0.93), $10-\mathrm{m}$ Sentinel-2 data (R2=0.91) \\
    \hline
    2022 \cite{li2022toward} & Automated machine learning for HSI analysis. Estimation of crop yield and biomass using AI. & AutoML framework for regression operations. Hyperspectral narrow-band vegetation indexes for modelling & R2 and NRMSE values for yield estimation: Wheat: (R2=0.96, NRMSE=0.12). R2 and NRMSE values for straw mass estimation: Wheat: (R2=0.96, NRMSE=0.12) \\
    \hline
    2022 \cite{li2022uav} & UAV-based hyperspectral and ensemble machine learning for predicting winter wheat yield. Spectral indices and machine learning models used for prediction. & Recursive feature elimination (RFE), Boruta feature selection, Pearson correlation coefficient (PCC) & The SVM yield prediction model performed the best (R2=0.73). (R2=0.78) for the yield prediction model based on Boruta's \\
    \hline
    2022 \cite{fei2022application} & Investigates the use of a multi-layer neural network and hyperspectral reflectance in a genome-wide association research for grain yield in bread wheat & Multi-layer neural network regression. & Multi-layer neural network model at the MGF stage: (R²=0.76), EGF stage: (R²=0.69), LGF stage: (R²=0.65), Flowering stage: (R²=0.68). Linear regression: (R²=0.87-0.94) \\
    \hline
    2023 \cite{liu2023estimation} & This research focuses on estimating winter wheat yield using UAV-based imagery. Multiple regression model combining growth stages for accurate estimation. & Simple linear regression model based on single growth stages; Multiple linear regression model combining multiple growth stages & (r=0.84; RMSE=0.69 t/ha) \\
    \hline
    2023 \cite{yang2023rapid} & Evaluate the chlorophyll content of wheat under drought stress conditions using HSI, & Random Forest Regression (RFR) & (R²=0.61; RMSE=4.439 t/ha; RE=7.35\%) \\
    \hline
    2024 \cite{zhang2024comparison} & Improve the accuracy and transferability of wheat yield estimation models by integrating various sensor data & LSTM with multi-head self-attention (LSTM-MH-SA) and joint Distribution Adaptation (JDA). & LSTM-MH-SA: (R²=0.2958); JDA: (R²= 0.81) \\
    \hline
    2024 \cite{zhao2024spatial} & Enhance the accuracy and robustness of in-situ leaf chlorophyll content (LCC) estimation & 3DCNN-LSTM  & (RP²=0.96) \\
    \hline
    2024 \cite{liu2024estimation} & Develop a more accurate wheat aboveground biomass (AGB) estimation model & Phenological information and vegetation indices (VIs) with the Lasso model (FIWheat-AGB)   & FIWheat-AGB: (R²=0.91, RMSE=2.11 t/ha, MAE=1 t/ha) \\
    \hline
    \end{tabular}
    \label{tab:Table 10}
\end{table}


\section{Discussion}
\label{sec:Section 7}

The integration of deep learning with hyperspectral imaging has significantly advanced wheat crop analysis, enabling accurate classification, disease detection, nutrient estimation, and yield prediction. While HSI surpasses traditional crop monitoring methods in precision and automation, classical detection techniques remain relevant due to their reliability, particularly in cases where HSI's high cost and computational demands pose challenges. The inaccuracies in manual analysis, variations in instrumentation, and limitations of chemical-based detection methods have accelerated the transition toward data-driven approaches, where DL plays a crucial role in extracting meaningful insights from high-dimensional spectral data. Consequently, various learning paradigms have been explored for wheat crop analysis, each suited to specific scenarios and challenges.

Supervised learning is the most widely investigated approach in wheat crop analysis due to its ability to learn from labeled datasets for precise predictions. CNNs have been the most commonly adopted models, effectively capturing spatial features in both 2D and 3D forms. However, their limited ability to model long-range spectral dependencies has led to the adoption of more advanced architectures. The introduction of transformers has revolutionized hyperspectral classification through the self-attention mechanism. More recently, Mamba models have emerged as a promising alternative, achieving state-of-the-art performance while maintaining computational efficiency. Despite these advancements, the reliance on large-scale labeled datasets remains a challenge.

Semi-supervised learning, on the other hand, has been instrumental in addressing the scarcity of labeled hyperspectral datasets. Pseudo-labeling techniques enable models to refine predictions on unlabeled data, while GANs have been particularly useful for synthetic data generation to balance dataset distributions. However, pseudo-labeling remains prone to label noise, and generative models often struggle to fully capture real-world spectral variability, limiting their effectiveness in certain applications.

In addition to supervised and semi-supervised learning, unsupervised learning has gained traction, particularly for hyperspectral data compression, feature extraction, and spectral unmixing. Stacked autoencoders and deep belief networks have been valuable for dimensionality reduction, improving model interpretability and computational efficiency. However, their application in wheat crop analysis remains limited due to challenges in defining meaningful feature representations without labeled data. More recently, diffusion models have shown promise in denoising and classification tasks, offering potential for further research.

From an application perspective, DL has demonstrated significant advancements across various tasks in wheat crop analysis. In wheat classification, CNNs, RNNs, DBNs, and SAEs have been widely used. More recently, transformers and Mamba models have been adopted, defining the state-of-the-art performance. In disease detection, CNNs remain the most widely used approach, though GANs have enhanced performance by generating synthetic hyperspectral samples. For nutrient estimation, CNNs have been applied to analyze spectral reflectance data, but future research should explore recurrent models (e.g., lLSTMs) for tracking nutrient variations over time. Similarly, in yield estimation, LSTMs have shown promise in modeling sequential crop growth patterns, while CNN-based regression models have been effective for biomass and grain yield prediction. However, environmental factors such as soil variation and climate conditions continue to challenge model generalization, highlighting the need for hybrid deep learning and agronomic models to improve real-world applicability.

\section{Challenges and Future Directions}
\label{sec:Section 8}
Despite the highlighted advancements in the previous sections, several challenges require further attention to fully harness deep learning for the hyperspectral analysis of wheat crops. One of the most pressing issues is the complexity and cost of HSI data acquisition. The high-dimensional nature of hyperspectral data requires substantial computational power, making real-time agricultural monitoring difficult. Although deep learning models such as CNNs, transformers, and Mamba have achieved impressive results, their computational demands remain a major barrier to widespread adoption. Future research should focus on lightweight architectures optimized for UAVs, edge computing, and mobile sensors to make real-time HSI analysis more accessible for farmers and agronomists.

Another key limitation is the lower spatial resolution of HSI compared to multispectral or RGB imaging, which makes it difficult to capture fine structural details in crops. Integrating super-resolution techniques and spectral-spatial fusion models can improve interpretability and enhance performance in applications such as disease detection and nutrient monitoring.

Additionally, the lack of publicly available, large, and diverse hyperspectral datasets significantly limits deep learning research in certain wheat crop analysis tasks. While classification has received more attention, datasets for disease detection, nutrient estimation, and yield prediction remain scarce or are often proprietary, restricting access to researchers. This shortage prevents the full exploration of recent deep learning advancements, such as transformers, Mamba models, and generative models, in these areas. Without access to sufficient labeled data, models struggle to generalize across diverse wheat-growing conditions, limiting their practical applicability.

Beyond these challenges, several promising research directions can enhance deep learning-based hyperspectral wheat crop analysis, including meta-learning and automatic hyperparameter and architecture optimization. Meta-learning enables models to generalize across datasets and remains underexplored in HSI despite its potential to improve adaptability in different agricultural environments. Automatic hyperparameter and architecture optimization techniques, such as genetic algorithms and evolutionary computing, can further enhance model performance.

Another important research direction is to identify and reduce redundancy in hyperspectral data cubes. The presence of redundant spectral bands introduces noise and inefficiencies, affecting classification accuracy. Developing automatic feature selection protocols can improve model efficiency, while designing compact and cost-effective HSI setups will accelerate practical adoption in agriculture.

Advancements in self-supervised learning and contrastive learning hold promise for reducing reliance on expensive labeled datasets by enabling models to learn from unlabeled data. The fusion of CNNs and transformers has already yielded promising results in hyperspectral classification, and as research progresses, their applications in agriculture will continue to expand. Additionally, Mamba models' ability to process large hyperspectral datasets efficiently will play a crucial role in making HSI analysis more scalable and feasible for real-world agricultural applications.

Deep learning has revolutionized hyperspectral wheat crop analysis, significantly improving the accuracy of classification, disease detection, nutrient monitoring, and yield prediction. However, challenges related to computational demands, dataset availability, and model interpretability need to be addressed to ensure real-world scalability. Future research should focus on developing lightweight, energy-efficient models, expanding labeled datasets, integrating Explainable AI techniques, and exploring hybrid learning approaches. By overcoming these limitations, AI-driven hyperspectral analysis will continue to evolve, offering transformative solutions for precision agriculture and sustainable wheat production.

\section{Conclusion}
\label{sec:Section 9}

This survey explores the transformative role of deep learning  in hyperspectral wheat crop analysis, highlighting its potential to enhance precision agriculture through non-destructive, data-driven insights. We categorize existing DL approaches into supervised, semi-supervised, and unsupervised learning, providing a comprehensive taxonomy that offers valuable guidance for both researchers and practitioners. By analyzing and comparing state-of-the-art methods, we highlight their effectiveness in classification, disease detection, nutrient estimation, and yield prediction, while also identifying opportunities for further advancements in this domain.


Despite the advancements in deep learning for wheat crop analysis using HSI, several challenges hinder the broader adoption of deep learning for hyperspectral wheat crop analysis. Limited public datasets remain a major bottleneck, restricting the generalization of DL models across diverse agricultural conditions. Many critical tasks, including disease detection, nutrient estimation, and yield prediction, suffer from a lack of openly available labeled hyperspectral datasets, which in turn limits the exploration and application of advanced deep learning models, such as transformers, Mamba models, and generative models. Additionally, the high computational demands of these models pose challenges for real-time deployment, particularly in resource-constrained agricultural settings. The reliance on specialized hardware, such as UAV-based and edge-computing solutions, further complicates large-scale adoption. Overcoming these challenges is crucial for the seamless integration of DL-powered HSI systems into precision agriculture.

To address these limitations, future research should focus on developing computationally efficient architectures, such as Mamba models, to accelerate hyperspectral data processing. Additionally, self-supervised learning, transfer learning, and meta-learning offer promising solutions to reduce dependency on large labeled datasets while enhancing model adaptability.  By fostering dataset expansion, model optimization, and interdisciplinary collaboration, AI-driven hyperspectral imaging will continue to drive innovation in wheat crop analysis, contributing to global food security and sustainable farming.



\begin{thebibliography}{100}
\providecommand{\url}[1]{#1}
\csname url@samestyle\endcsname
\providecommand{\newblock}{\relax}
\providecommand{\bibinfo}[2]{#2}
\providecommand{\BIBentrySTDinterwordspacing}{\spaceskip=0pt\relax}
\providecommand{\BIBentryALTinterwordstretchfactor}{4}
\providecommand{\BIBentryALTinterwordspacing}{\spaceskip=\fontdimen2\font plus
\BIBentryALTinterwordstretchfactor\fontdimen3\font minus \fontdimen4\font\relax}
\providecommand{\BIBforeignlanguage}[2]{{%
\expandafter\ifx\csname l@#1\endcsname\relax
\typeout{** WARNING: IEEEtran.bst: No hyphenation pattern has been}%
\typeout{** loaded for the language `#1'. Using the pattern for}%
\typeout{** the default language instead.}%
\else
\language=\csname l@#1\endcsname
\fi
#2}}
\providecommand{\BIBdecl}{\relax}
\BIBdecl

\bibitem{shiferaw2013crops}
B.~Shiferaw, M.~Smale, H.-J. Braun, E.~Duveiller, M.~Reynolds, and G.~Muricho, ``Crops that feed the world 10. past successes and future challenges to the role played by wheat in global food security,'' \emph{Food Security}, vol.~5, pp. 291--317, 2013.

\bibitem{acevedo2018role}
M.~Acevedo, J.~D. Zurn, G.~Molero, P.~Singh, X.~He, M.~Aoun, P.~Juliana, H.~Bockleman, M.~Bonman, M.~El-Sohl \emph{et~al.}, ``The role of wheat in global food security,'' in \emph{Agricultural development and sustainable intensification}.\hskip 1em plus 0.5em minus 0.4em\relax Routledge, 2018, pp. 81--110.

\bibitem{in1}
V.~Gonzalez-Dugo, P.~Hernandez, I.~Solis, and P.~J. Zarco-Tejada, ``Using high-resolution hyperspectral and thermal airborne imagery to assess physiological condition in the context of wheat phenotyping,'' \emph{Remote Sensing}, vol.~7, no.~10, pp. 13\,586--13\,605, 2015.

\bibitem{in2}
I.~Mariotto, P.~S. Thenkabail, A.~Huete, E.~T. Slonecker, and A.~Platonov, ``Hyperspectral versus multispectral crop-productivity modeling and type discrimination for the hyspiri mission,'' \emph{Remote Sensing of Environment}, vol. 139, pp. 291--305, 2013.

\bibitem{in3}
M.~Marshall and P.~Thenkabail, ``Advantage of hyperspectral eo-1 hyperion over multispectral ikonos, geoeye-1, worldview-2, landsat etm+, and modis vegetation indices in crop biomass estimation,'' \emph{ISPRS Journal of Photogrammetry and Remote Sensing}, vol. 108, pp. 205--218, 2015.

\bibitem{in4}
V.~Lodhi, D.~Chakravarty, and P.~Mitra, ``Hyperspectral imaging system: Development aspects and recent trends,'' \emph{Sensing and Imaging}, vol.~20, pp. 1--24, 2019.

\bibitem{ang2021big}
K.~L.-M. Ang and J.~K.~P. Seng, ``Big data and machine learning with hyperspectral information in agriculture,'' \emph{IEEE Access}, vol.~9, pp. 36\,699--36\,718, 2021.

\bibitem{khan2022systematic}
A.~Khan, A.~D. Vibhute, S.~Mali, and C.~Patil, ``A systematic review on hyperspectral imaging technology with a machine and deep learning methodology for agricultural applications,'' \emph{Ecological Informatics}, vol.~69, p. 101678, 2022.

\bibitem{li2022uav}
Z.~Li, Z.~Chen, Q.~Cheng, F.~Duan, R.~Sui, X.~Huang, and H.~Xu, ``Uav-based hyperspectral and ensemble machine learning for predicting yield in winter wheat,'' \emph{Agronomy}, vol.~12, no.~1, p. 202, 2022.

\bibitem{tejasree2024extensive}
G.~Tejasree and L.~Agilandeeswari, ``An extensive review of hyperspectral image classification and prediction: techniques and challenges,'' \emph{Multimedia Tools and Applications}, pp. 1--98, 2024.

\bibitem{wang2021review}
C.~Wang, B.~Liu, L.~Liu, Y.~Zhu, J.~Hou, P.~Liu, and X.~Li, ``A review of deep learning used in the hyperspectral image analysis for agriculture,'' \emph{Artificial Intelligence Review}, vol.~54, no.~7, pp. 5205--5253, 2021.

\bibitem{khan2022systematicc}
A.~Khan, A.~D. Vibhute, S.~Mali, and C.~Patil, ``A systematic review on hyperspectral imaging technology with a machine and deep learning methodology for agricultural applications,'' \emph{Ecological Informatics}, vol.~69, p. 101678, 2022.

\bibitem{kuswidiyanto2022plant}
L.~W. Kuswidiyanto, H.-H. Noh, and X.~Han, ``Plant disease diagnosis using deep learning based on aerial hyperspectral images: A review,'' \emph{Remote Sensing}, vol.~14, no.~23, p. 6031, 2022.

\bibitem{abdelkrim2024hyperspectral}
O.~Abdelkrim \emph{et~al.}, ``Hyperspectral imaging using deep learning in wheat diseases,'' in \emph{2024 8th International Conference on Image and Signal Processing and their Applications (ISPA)}.\hskip 1em plus 0.5em minus 0.4em\relax IEEE, 2024, pp. 1--8.

\bibitem{kaur2024hyperspectral}
S.~Kaur, V.~G. Kakani, B.~Carver, D.~Jarquin, and A.~Singh, ``Hyperspectral imaging combined with machine learning for high-throughput phenotyping in winter wheat,'' \emph{The Plant Phenome Journal}, vol.~7, no.~1, p. e20111, 2024.

\bibitem{sharma2024nondestructive}
A.~Sharma, T.~Singh, and N.~M. Garg, ``Nondestructive identification of wheat seed variety and geographical origin using near-infrared hyperspectral imagery and deep learning,'' \emph{Journal of Chemometrics}, p. e3585, 2024.

\bibitem{caballero2019hyperspectral}
D.~Caballero, R.~Calvini, and J.~M. Amigo, ``Hyperspectral imaging in crop fields: precision agriculture,'' in \emph{Data handling in science and technology}.\hskip 1em plus 0.5em minus 0.4em\relax Elsevier, 2019, vol.~32, pp. 453--473.

\bibitem{example}
``Hyperspectral imaging,'' \url{https://www.eoportal.org/other-space-activities/hyperspectral-imaging#ghost}, 2023, accessed on January 30, 2024.

\bibitem{wheat}
D.~R.~K. Marja~Haagsma, Christina H.~Hagerty and J.~S. Selker, ``Detection of soil-borne wheat mosaic virus using hyperspectral imaging: from lab to field scans and from hyperspectral to multispectral data,'' \emph{SpringerLink}, vol.~24, 2023.

\bibitem{pict}
nasa.gov, ``Aviris data - aviris moffett field image cube,'' \url{https://aviris.jpl.nasa.gov/data/image_cube.html}, 2022.

\bibitem{adao2017hyperspectral}
T.~Ad{\~a}o, J.~Hru{\v{s}}ka, L.~P{\'a}dua, J.~Bessa, E.~Peres, R.~Morais, and J.~J. Sousa, ``Hyperspectral imaging: A review on uav-based sensors, data processing and applications for agriculture and forestry,'' \emph{Remote sensing}, vol.~9, no.~11, p. 1110, 2017.

\bibitem{lu2020recent}
B.~Lu, P.~D. Dao, J.~Liu, Y.~He, and J.~Shang, ``Recent advances of hyperspectral imaging technology and applications in agriculture,'' \emph{Remote Sensing}, vol.~12, no.~16, p. 2659, 2020.

\bibitem{lodhi2018hyperspectral}
V.~Lodhi, D.~Chakravarty, and P.~Mitra, ``Hyperspectral imaging for earth observation: Platforms and instruments,'' \emph{Journal of the Indian Institute of Science}, vol.~98, pp. 429--443, 2018.

\bibitem{vorovencii2010hyperspectral}
I.~Vorovencii, ``The hyperspectral sensors used in satellite and aerial remote sensing,'' \emph{Bulletin of the Transilvania University of Brasov. Series II: Forestry• Wood Industry• Agricultural Food Engineering}, pp. 51--56, 2010.

\bibitem{habib2016improving}
A.~Habib, W.~Xiong, F.~He, H.~L. Yang, and M.~Crawford, ``Improving orthorectification of uav-based push-broom scanner imagery using derived orthophotos from frame cameras,'' \emph{IEEE Journal of Selected Topics in Applied Earth Observations and Remote Sensing}, vol.~10, no.~1, pp. 262--276, 2016.

\bibitem{transon2018survey}
J.~Transon, R.~d’Andrimont, A.~Maugnard, and P.~Defourny, ``Survey of hyperspectral earth observation applications from space in the sentinel-2 context,'' \emph{Remote Sensing}, vol.~10, no.~2, p. 157, 2018.

\bibitem{lodhi2018hyperspectrall}
V.~Lodhi, D.~Chakravarty, and P.~Mitra, ``Hyperspectral imaging for earth observation: Platforms and instruments,'' \emph{Journal of the Indian Institute of Science}, vol.~98, pp. 429--443, 2018.

\bibitem{jia2020overview}
X.~Jia, S.~Li, S.~Ke, and B.~Hu, ``Overview of spaceborne hyperspectral imagers and the research progress in bathymetric maps,'' in \emph{Second Target Recognition and Artificial Intelligence Summit Forum}, vol. 11427.\hskip 1em plus 0.5em minus 0.4em\relax SPIE, 2020, pp. 118--124.

\bibitem{ranjan2012assessment}
R.~Ranjan, U.~K. Chopra, R.~N. Sahoo, A.~K. Singh, and S.~Pradhan, ``Assessment of plant nitrogen stress in wheat (triticum aestivum l.) through hyperspectral indices,'' \emph{International Journal of Remote Sensing}, vol.~33, no.~20, pp. 6342--6360, 2012.

\bibitem{tao2020estimation}
H.~Tao, H.~Feng, L.~Xu, M.~Miao, G.~Yang, X.~Yang, and L.~Fan, ``Estimation of the yield and plant height of winter wheat using uav-based hyperspectral images,'' \emph{Sensors}, vol.~20, no.~4, p. 1231, 2020.

\bibitem{gonzalez2015using}
V.~Gonzalez-Dugo, P.~Hernandez, I.~Solis, and P.~J. Zarco-Tejada, ``Using high-resolution hyperspectral and thermal airborne imagery to assess physiological condition in the context of wheat phenotyping,'' \emph{Remote Sensing}, vol.~7, no.~10, pp. 13\,586--13\,605, 2015.

\bibitem{montesinos2017predicting}
O.~A. Montesinos-L{\'o}pez, A.~Montesinos-L{\'o}pez, J.~Crossa, G.~de~Los~Campos, G.~Alvarado, M.~Suchismita, J.~Rutkoski, L.~Gonz{\'a}lez-P{\'e}rez, and J.~Burgue{\~n}o, ``Predicting grain yield using canopy hyperspectral reflectance in wheat breeding data,'' \emph{Plant methods}, vol.~13, pp. 1--23, 2017.

\bibitem{chang2013hyperspectral}
C.-I. Chang, \emph{Hyperspectral data processing: algorithm design and analysis}.\hskip 1em plus 0.5em minus 0.4em\relax John Wiley \& Sons, 2013.

\bibitem{sun2021spassa}
G.~Sun, H.~Fu, J.~Ren, A.~Zhang, J.~Zabalza, X.~Jia, and H.~Zhao, ``Spassa: Superpixelwise adaptive ssa for unsupervised spatial--spectral feature extraction in hyperspectral image,'' \emph{IEEE Transactions on Cybernetics}, vol.~52, no.~7, pp. 6158--6169, 2021.

\bibitem{ngadi2010hyperspectral}
M.~O. Ngadi and L.~Liu, ``Hyperspectral image processing techniques,'' in \emph{Hyperspectral imaging for food quality analysis and control}.\hskip 1em plus 0.5em minus 0.4em\relax Elsevier, 2010, pp. 99--127.

\bibitem{site2}
``hyperspectral-image-processing,'' \url{https://www.sciencedirect.com/topics/computer-science/hyperspectral-image-processing}, 2023.

\bibitem{site3}
``Basic hyperspectral analysis tutorial,'' \url{https://www.nv5geospatialsoftware.com/docs/HyperspectralAnalysisTutorial.html}, 2023.

\bibitem{zhao2019featureexplorer}
J.~Zhao, M.~Karimzadeh, A.~Masjedi, T.~Wang, X.~Zhang, M.~M. Crawford, and D.~S. Ebert, ``Featureexplorer: Interactive feature selection and exploration of regression models for hyperspectral images,'' in \emph{2019 IEEE Visualization Conference (VIS)}.\hskip 1em plus 0.5em minus 0.4em\relax IEEE, 2019, pp. 161--165.

\bibitem{xu2019superpixel}
H.~Xu, H.~Zhang, W.~He, and L.~Zhang, ``Superpixel-based spatial-spectral dimension reduction for hyperspectral imagery classification,'' \emph{Neurocomputing}, vol. 360, pp. 138--150, 2019.

\bibitem{prasad2020hyperspectral}
S.~Prasad and J.~Chanussot, \emph{Hyperspectral image analysis: advances in machine learning and signal processing}.\hskip 1em plus 0.5em minus 0.4em\relax Springer Nature, 2020.

\bibitem{hy}
E.~Taghinezhad, A.~Szumny, and A.~Figiel, ``The application of hyperspectral imaging technologies for the prediction and measurement of the moisture content of various agricultural crops during the drying process,'' \emph{Molecules}, vol.~28, no.~7, p. 2930, 2023.

\bibitem{site5}
B.~A. M~Graña, MA~Veganzons, ``Hyperspectral remote sensing scenes,'' \url{https://www.ehu.eus/ccwintco/index.php/Hyperspectral_Remote_Sensing_Scenes#Indian_Pines.}, 1992/1998.

\bibitem{site6}
``Global hyperspectral imaging spectral-library of agricultural crops for conterminous united states v001,'' \url{https://catalog.data.gov/dataset/global-hyperspectral-imaging-spectral-library-of-agricultural-crops-for-conterminous-unite}, December 7, 2023.

\bibitem{site9}
E.~R. Haagsma Marja~Hagerty, Christina H. Kroese Duncan R.~Jachens, ``Hyperspectral reflectance data of wheat in situ and lab scans infected with soilborne wheat mosaic virus,'' \url{https://ir.library.oregonstate.edu/concern/datasets/z316q855z}, 2019.

\bibitem{site7}
H.~Liu, ``The promise of hyperspectral imaging for the early detection of crown rot in wheat,'' \url{https://adelaide.figshare.com/articles/dataset/The_Promise_of_Hyperspectral_Imaging_for_the_Early_Detection_of_Crown_Rot_in_Wheat/17075138}, 2021.

\bibitem{site8}
K.~Dhakal, ``Wheathyperspectral,'' \url{https://github.com/LiLabAtVT/WheatHyperSpectral}, 2021.

\bibitem{site10}
C.~Moghimi, Ali~Yang, ``Hyperspectral image dataset for salt stress phenotyping of wheat,'' \url{https://conservancy.umn.edu/handle/11299/195720}, April 13, 2018.

\bibitem{privi2}
E.~Bauriegel, A.~Giebel, M.~Geyer, U.~Schmidt, and W.~Herppich, ``Early detection of fusarium infection in wheat using hyper-spectral imaging,'' \emph{Computers and electronics in agriculture}, vol.~75, no.~2, pp. 304--312, 2011.

\bibitem{privi5}
H.~Ma, W.~Huang, Y.~Jing, S.~Pignatti, G.~Laneve, Y.~Dong, H.~Ye, L.~Liu, A.~Guo, and J.~Jiang, ``Identification of fusarium head blight in winter wheat ears using continuous wavelet analysis,'' \emph{Sensors}, vol.~20, no.~1, p.~20, 2019.

\bibitem{pri3}
Q.~Zheng, W.~Huang, X.~Cui, Y.~Dong, Y.~Shi, H.~Ma, and L.~Liu, ``Identification of wheat yellow rust using optimal three-band spectral indices in different growth stages,'' \emph{Sensors}, vol.~19, no.~1, p.~35, 2018.

\bibitem{pri4}
D.~Bohnenkamp, J.~Behmann, and A.-K. Mahlein, ``In-field detection of yellow rust in wheat on the ground canopy and uav scale,'' \emph{Remote Sensing}, vol.~11, no.~21, p. 2495, 2019.

\bibitem{pri5}
X.~Zhang, L.~Han, Y.~Dong, Y.~Shi, W.~Huang, L.~Han, P.~Gonz{\'a}lez-Moreno, H.~Ma, H.~Ye, and T.~Sobeih, ``A deep learning-based approach for automated yellow rust disease detection from high-resolution hyperspectral uav images,'' \emph{Remote Sensing}, vol.~11, no.~13, p. 1554, 2019.

\bibitem{figueroa2018review}
M.~Figueroa, K.~E. Hammond-Kosack, and P.~S. Solomon, ``A review of wheat diseases—a field perspective,'' \emph{Molecular plant pathology}, vol.~19, no.~6, pp. 1523--1536, 2018.

\bibitem{privi3}
J.~G. Barbedo, C.~S. Tibola, and J.~M. Fernandes, ``Detecting fusarium head blight in wheat kernels using hyperspectral imaging,'' \emph{Biosystems Engineering}, vol. 131, pp. 65--76, 2015.

\bibitem{privi4}
R.~L. Whetton, K.~L. Hassall, T.~W. Waine, and A.~M. Mouazen, ``Hyperspectral measurements of yellow rust and fusarium head blight in cereal crops: Part 1: Laboratory study,'' \emph{Biosystems Engineering}, vol. 166, pp. 101--115, 2018.

\bibitem{privi1}
N.~Zhang, Y.~Pan, H.~Feng, X.~Zhao, X.~Yang, C.~Ding, and G.~Yang, ``Development of fusarium head blight classification index using hyperspectral microscopy images of winter wheat spikelets,'' \emph{Biosystems Engineering}, vol. 186, pp. 83--99, 2019.

\bibitem{privi6}
D.~Zhang, Q.~Wang, F.~Lin, X.~Yin, C.~Gu, and H.~Qiao, ``Development and evaluation of a new spectral disease index to detect wheat fusarium head blight using hyperspectral imaging,'' \emph{Sensors}, vol.~20, no.~8, p. 2260, 2020.

\bibitem{privi7}
L.~Huang, H.~Zhang, C.~Ruan, W.~Huang, T.~Hu, and J.~Zhao, ``Detection of scab in wheat ears using in situ hyperspectral data and support vector machine optimized by genetic algorithm,'' \emph{International Journal of Agricultural and Biological Engineering}, vol.~13, no.~2, pp. 182--188, 2020.

\bibitem{pri1}
W.~Huang, D.~W. Lamb, Z.~Niu, Y.~Zhang, L.~Liu, and J.~Wang, ``Identification of yellow rust in wheat using in-situ spectral reflectance measurements and airborne hyperspectral imaging,'' \emph{Precision Agriculture}, vol.~8, pp. 187--197, 2007.

\bibitem{pri2}
G.~Krishna, R.~Sahoo, S.~Pargal, V.~Gupta, P.~Sinha, S.~Bhagat, M.~Saharan, R.~Singh, and C.~Chattopadhyay, ``Assessing wheat yellow rust disease through hyperspectral remote sensing,'' \emph{The international archives of the photogrammetry, remote sensing and spatial information sciences}, vol.~40, pp. 1413--1416, 2014.

\bibitem{pri6}
A.~Guo, W.~Huang, H.~Ye, Y.~Dong, H.~Ma, Y.~Ren, and C.~Ruan, ``Identification of wheat yellow rust using spectral and texture features of hyperspectral images,'' \emph{Remote Sensing}, vol.~12, no.~9, p. 1419, 2020.

\bibitem{paoletti2019deep}
M.~Paoletti, J.~Haut, J.~Plaza, and A.~Plaza, ``Deep learning classifiers for hyperspectral imaging: A review,'' \emph{ISPRS Journal of Photogrammetry and Remote Sensing}, vol. 158, pp. 279--317, 2019.

\bibitem{signoroni2019deep}
A.~Signoroni, M.~Savardi, A.~Baronio, and S.~Benini, ``Deep learning meets hyperspectral image analysis: A multidisciplinary review,'' \emph{Journal of Imaging}, vol.~5, no.~5, p.~52, 2019.

\bibitem{sadeghi2021neural}
P.~Sadeghi-Tehran, N.~Virlet, and M.~J. Hawkesford, ``A neural network method for classification of sunlit and shaded components of wheat canopies in the field using high-resolution hyperspectral imagery,'' \emph{Remote Sensing}, vol.~13, no.~5, p. 898, 2021.

\bibitem{dhakal2023machine}
K.~Dhakal, U.~Sivaramakrishnan, X.~Zhang, K.~Belay, J.~Oakes, X.~Wei, and S.~Li, ``Machine learning analysis of hyperspectral images of damaged wheat kernels,'' \emph{Sensors}, vol.~23, no.~7, p. 3523, 2023.

\bibitem{zhong2017spectral}
Z.~Zhong, J.~Li, Z.~Luo, and M.~Chapman, ``Spectral--spatial residual network for hyperspectral image classification: A 3-d deep learning framework,'' \emph{IEEE Transactions on Geoscience and Remote Sensing}, vol.~56, no.~2, pp. 847--858, 2017.

\bibitem{chen2016deep}
Y.~Chen, H.~Jiang, C.~Li, X.~Jia, and P.~Ghamisi, ``Deep feature extraction and classification of hyperspectral images based on convolutional neural networks,'' \emph{IEEE transactions on geoscience and remote sensing}, vol.~54, no.~10, pp. 6232--6251, 2016.

\bibitem{yang2018hyperspectral}
X.~Yang, Y.~Ye, X.~Li, R.~Y. Lau, X.~Zhang, and X.~Huang, ``Hyperspectral image classification with deep learning models,'' \emph{IEEE Transactions on Geoscience and Remote Sensing}, vol.~56, no.~9, pp. 5408--5423, 2018.

\bibitem{bing2019deep}
L.~Bing, Y.~Xuchu, Z.~Pengqiang, and T.~Xiong, ``Deep 3d convolutional network combined with spatial-spectral features for hyperspectral image classification,'' \emph{Acta Geodaetica et Cartographica Sinica}, vol.~48, no.~1, p.~53, 2019.

\bibitem{roy2019hybridsn}
S.~K. Roy, G.~Krishna, S.~R. Dubey, and B.~B. Chaudhuri, ``Hybridsn: Exploring 3-d--2-d cnn feature hierarchy for hyperspectral image classification,'' \emph{IEEE Geoscience and Remote Sensing Letters}, vol.~17, no.~2, pp. 277--281, 2019.

\bibitem{cnn1}
M.~Ashfaq and I.~Khan, ``Crop yield prediction of wheat using cnn and remote sensing multispectral imagery, a hybrid approach,'' 2023.

\bibitem{song2018hyperspectral}
W.~Song, S.~Li, L.~Fang, and T.~Lu, ``Hyperspectral image classification with deep feature fusion network,'' \emph{IEEE Transactions on Geoscience and Remote Sensing}, vol.~56, no.~6, pp. 3173--3184, 2018.

\bibitem{rnn1}
N.~Kussul, M.~Lavreniuk, and L.~Shumilo, ``Deep recurrent neural network for crop classification task based on sentinel-1 and sentinel-2 imagery,'' in \emph{IGARSS 2020-2020 IEEE International Geoscience and Remote Sensing Symposium}.\hskip 1em plus 0.5em minus 0.4em\relax IEEE, 2020, pp. 6914--6917.

\bibitem{venkatesan2019hyperspectral}
R.~Venkatesan and S.~Prabu, ``Hyperspectral image features classification using deep learning recurrent neural networks,'' \emph{Journal of medical systems}, vol.~43, no.~7, p. 216, 2019.

\bibitem{luo2018shorten}
H.~Luo, ``Shorten spatial-spectral rnn with parallel-gru for hyperspectral image classification,'' \emph{arXiv preprint arXiv:1810.12563}, 2018.

\bibitem{dang2023double}
L.~Dang, L.~Weng, Y.~Hou, X.~Zuo, and Y.~Liu, ``Double-branch feature fusion transformer for hyperspectral image classification,'' \emph{Scientific Reports}, vol.~13, no.~1, p. 272, 2023.

\bibitem{tran1}
N.~Sigger, Q.-T. Vien, S.~V. Nguyen, G.~Tozzi, and T.~T. Nguyen, ``Unveiling the potential of diffusion model-based framework with transformer for hyperspectral image classification,'' \emph{Scientific Reports}, vol.~14, no.~1, p. 8438, 2024.

\bibitem{tran2}
S.~Liu, C.~Yin, and H.~Zhang, ``Cesa-mcformer: An efficient transformer network for hyperspectral image classification by eliminating redundant information,'' \emph{Sensors}, vol.~24, no.~4, p. 1187, 2024.

\bibitem{tran3}
J.~Xie, J.~Hua, S.~Chen, P.~Wu, P.~Gao, D.~Sun, Z.~Lyu, S.~Lyu, X.~Xue, and J.~Lu, ``Hypersformer: A transformer-based end-to-end hyperspectral image classification method for crop classification,'' \emph{Remote Sensing}, vol.~15, no.~14, p. 3491, 2023.

\bibitem{sheng2024dualmamba}
J.~Sheng, J.~Zhou, J.~Wang, P.~Ye, and J.~Fan, ``Dualmamba: A lightweight spectral-spatial mamba-convolution network for hyperspectral image classification,'' \emph{arXiv preprint arXiv:2406.07050}, 2024.

\bibitem{yang2024hsimamba}
J.~X. Yang, J.~Zhou, J.~Wang, H.~Tian, and A.~W.~C. Liew, ``Hsimamba: Hyperpsectral imaging efficient feature learning with bidirectional state space for classification,'' \emph{arXiv preprint arXiv:2404.00272}, 2024.

\bibitem{zhou2024mamba}
W.~Zhou, S.-I. Kamata, H.~Wang, M.-S. Wong \emph{et~al.}, ``Mamba-in-mamba: Centralized mamba-cross-scan in tokenized mamba model for hyperspectral image classification,'' \emph{arXiv preprint arXiv:2405.12003}, 2024.

\bibitem{huang2024spectral}
L.~Huang, Y.~Chen, and X.~He, ``Spectral-spatial mamba for hyperspectral image classification,'' \emph{arXiv preprint arXiv:2404.18401}, 2024.

\bibitem{he20243dss}
Y.~He, B.~Tu, B.~Liu, J.~Li, and A.~Plaza, ``3dss-mamba: 3d-spectral-spatial mamba for hyperspectral image classification,'' \emph{arXiv preprint arXiv:2405.12487}, 2024.

\bibitem{yang2024graphmamba}
A.~Yang, M.~Li, Y.~Ding, L.~Fang, Y.~Cai, and Y.~He, ``Graphmamba: An efficient graph structure learning vision mamba for hyperspectral image classification,'' \emph{arXiv preprint arXiv:2407.08255}, 2024.

\bibitem{dbn1}
C.~Li, Y.~Wang, X.~Zhang, H.~Gao, Y.~Yang, and J.~Wang, ``Deep belief network for spectral--spatial classification of hyperspectral remote sensor data,'' \emph{Sensors}, vol.~19, no.~1, p. 204, 2019.

\bibitem{sellami2019spectra}
A.~Sellami and I.~Farah, ``Spectra-spatial graph-based deep restricted boltzmann networks for hyperspectral image classification,'' in \emph{2019 PhotonIcs \& Electromagnetics Research Symposium-Spring (PIERS-Spring)}.\hskip 1em plus 0.5em minus 0.4em\relax IEEE, 2019, pp. 1055--1062.

\bibitem{mughees2018multiple}
A.~Mughees and L.~Tao, ``Multiple deep-belief-network-based spectral-spatial classification of hyperspectral images,'' \emph{Tsinghua Science and Technology}, vol.~24, no.~2, pp. 183--194, 2018.

\bibitem{zhong2017learning}
P.~Zhong, Z.~Gong, S.~Li, and C.-B. Sch{\"o}nlieb, ``Learning to diversify deep belief networks for hyperspectral image classification,'' \emph{IEEE Transactions on Geoscience and Remote Sensing}, vol.~55, no.~6, pp. 3516--3530, 2017.

\bibitem{xing2016stacked}
C.~Xing, L.~Ma, and X.~Yang, ``Stacked denoise autoencoder based feature extraction and classification for hyperspectral images,'' \emph{Journal of Sensors}, vol. 2016, no.~1, p. 3632943, 2016.

\bibitem{bai2024two}
Y.~Bai, X.~Sun, Y.~Ji, W.~Fu, and J.~Zhang, ``Two-stage multi-dimensional convolutional stacked autoencoder network model for hyperspectral images classification,'' \emph{Multimedia Tools and Applications}, vol.~83, no.~8, pp. 23\,489--23\,508, 2024.

\bibitem{zhao2024enhancing}
J.~Zhao, J.~Zhang, H.~Huang, and J.~Zhang, ``Enhancing semi-supervised few-shot hyperspectral image classification via progressive sample selection,'' \emph{Remote Sensing}, vol.~16, no.~10, p. 1747, 2024.

\bibitem{wang2021improved}
Q.~Wang, M.~Chen, J.~Zhang, S.~Kang, and Y.~Wang, ``Improved active deep learning for semi-supervised classification of hyperspectral image,'' \emph{Remote Sensing}, vol.~14, no.~1, p. 171, 2021.

\bibitem{wu2017semi}
H.~Wu and S.~Prasad, ``Semi-supervised deep learning using pseudo labels for hyperspectral image classification,'' \emph{IEEE Transactions on Image Processing}, vol.~27, no.~3, pp. 1259--1270, 2017.

\bibitem{fang2018semi}
B.~Fang, Y.~Li, H.~Zhang, and J.~C.-W. Chan, ``Semi-supervised deep learning classification for hyperspectral image based on dual-strategy sample selection,'' \emph{Remote Sensing}, vol.~10, no.~4, p. 574, 2018.

\bibitem{zhang2020semi}
Z.~Zhang, ``Semi-supervised hyperspectral image classification algorithm based on graph embedding and discriminative spatial information,'' \emph{Microprocessors and Microsystems}, vol.~75, p. 103070, 2020.

\bibitem{gan3}
W.~Zhang, Z.~Li, G.~Li, P.~Zhuang, G.~Hou, Q.~Zhang, and C.~Li, ``Gacnet: Generate adversarial-driven cross-aware network for hyperspectral wheat variety identification,'' \emph{IEEE Transactions on Geoscience and Remote Sensing}, 2023.

\bibitem{he2017generative}
Z.~He, H.~Liu, Y.~Wang, and J.~Hu, ``Generative adversarial networks-based semi-supervised learning for hyperspectral image classification,'' \emph{Remote Sensing}, vol.~9, no.~10, p. 1042, 2017.

\bibitem{zhan2023semisupervised}
Y.~Zhan, Y.~Wang, and X.~Yu, ``Semisupervised hyperspectral image classification based on generative adversarial networks and spectral angle distance,'' \emph{Scientific Reports}, vol.~13, no.~1, p. 22019, 2023.

\bibitem{goodfellow2020generative}
I.~Goodfellow, J.~Pouget-Abadie, M.~Mirza, B.~Xu, D.~Warde-Farley, S.~Ozair, A.~Courville, and Y.~Bengio, ``Generative adversarial networks,'' \emph{Communications of the ACM}, vol.~63, no.~11, pp. 139--144, 2020.

\bibitem{chen2019hyperspectral}
F.~Chen, J.~Li, and D.~Yang, ``Hyperspectral image classification based on generative adversarial networks,'' \emph{Comput Eng Appl}, vol.~55, no.~22, pp. 172--179, 2019.

\bibitem{zhu2018generative}
L.~Zhu, Y.~Chen, P.~Ghamisi, and J.~A. Benediktsson, ``Generative adversarial networks for hyperspectral image classification,'' \emph{IEEE Transactions on Geoscience and Remote Sensing}, vol.~56, no.~9, pp. 5046--5063, 2018.

\bibitem{liu2020cascade}
X.~Liu, Y.~Qiao, Y.~Xiong, Z.~Cai, and P.~Liu, ``Cascade conditional generative adversarial nets for spatial-spectral hyperspectral sample generation,'' \emph{Science China Information Sciences}, vol.~63, pp. 1--16, 2020.

\bibitem{yang2019feature}
J.~Yang, Y.~Guo, and X.~Wang, ``Feature extraction of hyperspectral images based on deep boltzmann machine,'' \emph{IEEE Geoscience and Remote Sensing Letters}, vol.~17, no.~6, pp. 1077--1081, 2019.

\bibitem{li2022manifold}
Z.~Li, H.~Huang, Z.~Zhang, and G.~Shi, ``Manifold-based multi-deep belief network for feature extraction of hyperspectral image,'' \emph{Remote Sensing}, vol.~14, no.~6, p. 1484, 2022.

\bibitem{sae1}
K.~Liang, J.~Huang, R.~He, Q.~Wang, Y.~Chai, and M.~Shen, ``Comparison of vis-nir and swir hyperspectral imaging for the non-destructive detection of don levels in fusarium head blight wheat kernels and wheat flour,'' \emph{Infrared Physics \& Technology}, vol. 106, p. 103281, 2020.

\bibitem{mughees2016efficient}
A.~Mughees and L.~Tao, ``Efficient deep auto-encoder learning for the classification of hyperspectral images,'' in \emph{2016 international conference on virtual reality and visualization (ICVRV)}.\hskip 1em plus 0.5em minus 0.4em\relax IEEE, 2016, pp. 44--51.

\bibitem{afrin2024enhancing}
A.~Afrin, M.~R. Haque, and M.~Al~Mamun, ``Enhancing hyperspectral image compression through stacked autoencoder approach,'' in \emph{2024 6th International Conference on Electrical Engineering and Information \& Communication Technology (ICEEICT)}.\hskip 1em plus 0.5em minus 0.4em\relax IEEE, 2024, pp. 1372--1377.

\bibitem{deng2023noise}
L.~Deng, B.~Zhou, J.~Ying, and R.~Zhao, ``A noise estimation method for hyperspectral image based on stacked autoencoder,'' \emph{IEEE Access}, 2023.

\bibitem{cao2024two}
C.~Cao, W.~Song, H.~Xiang, H.~Yi, F.~Xiao, and X.~Gao, ``A two-stream stacked autoencoder with inter-class separability for bilinear hyperspectral unmixing,'' \emph{IEEE Transactions on Computational Imaging}, 2024.

\bibitem{pang2024hir}
L.~Pang, X.~Rui, L.~Cui, H.~Wang, D.~Meng, and X.~Cao, ``Hir-diff: Unsupervised hyperspectral image restoration via improved diffusion models,'' in \emph{Proceedings of the IEEE/CVF Conference on Computer Vision and Pattern Recognition}, 2024, pp. 3005--3014.

\bibitem{zhang2023diffucd}
X.~Zhang, S.~Tian, G.~Wang, H.~Zhou, and L.~Jiao, ``Diffucd: Unsupervised hyperspectral image change detection with semantic correlation diffusion model,'' \emph{arXiv preprint arXiv:2305.12410}, 2023.

\bibitem{polk2023unsupervised}
S.~L. Polk, K.~Cui, A.~H. Chan, D.~A. Coomes, R.~J. Plemmons, and J.~M. Murphy, ``Unsupervised diffusion and volume maximization-based clustering of hyperspectral images,'' \emph{Remote Sensing}, vol.~15, no.~4, p. 1053, 2023.

\bibitem{10264172}
Z.~Li, K.~Bi, Y.~Wang, Z.~Fang, and J.~Zhang, ``Supervised contrastive learning for open-set hyperspectral image classification,'' \emph{IEEE Geoscience and Remote Sensing Letters}, vol.~20, pp. 1--5, 2023.

\bibitem{citation}
\BIBentryALTinterwordspacing
V.~K.~S. Somenath~Bera and S.~C. Satapathy, ``Advances in hyperspectral image classification based on convolutional neural networks: A review,'' \emph{Tech Science Press (TSP)}, 2022. [Online]. Available: \url{https://www.techscience.com/CMES/v133n2/48965/html}
\BIBentrySTDinterwordspacing

\bibitem{li2017spectral}
Y.~Li, H.~Zhang, and Q.~Shen, ``Spectral--spatial classification of hyperspectral imagery with 3d convolutional neural network,'' \emph{Remote Sensing}, vol.~9, no.~1, p.~67, 2017.

\bibitem{fang2019deep}
L.~Fang, Z.~Liu, and W.~Song, ``Deep hashing neural networks for hyperspectral image feature extraction,'' \emph{IEEE Geoscience and Remote Sensing Letters}, vol.~16, no.~9, pp. 1412--1416, 2019.

\bibitem{lee2017going}
H.~Lee and H.~Kwon, ``Going deeper with contextual cnn for hyperspectral image classification,'' \emph{IEEE Transactions on Image Processing}, vol.~26, no.~10, pp. 4843--4855, 2017.

\bibitem{chung2014empirical}
J.~Chung, C.~Gulcehre, K.~Cho, and Y.~Bengio, ``Empirical evaluation of gated recurrent neural networks on sequence modeling,'' \emph{arXiv preprint arXiv:1412.3555}, 2014.

\bibitem{graves2012long}
A.~Graves and A.~Graves, ``Long short-term memory,'' \emph{Supervised sequence labelling with recurrent neural networks}, pp. 37--45, 2012.

\bibitem{mou2017deep}
L.~Mou, P.~Ghamisi, and X.~X. Zhu, ``Deep recurrent neural networks for hyperspectral image classification,'' \emph{IEEE transactions on geoscience and remote sensing}, vol.~55, no.~7, pp. 3639--3655, 2017.

\bibitem{rnn3}
E.~Ndikumana, D.~Ho~Tong~Minh, N.~Baghdadi, D.~Courault, and L.~Hossard, ``Deep recurrent neural network for agricultural classification using multitemporal sar sentinel-1 for camargue, france,'' \emph{Remote Sensing}, vol.~10, no.~8, p. 1217, 2018.

\bibitem{salmela2021predicting}
L.~Salmela, N.~Tsipinakis, A.~Foi, C.~Billet, J.~M. Dudley, and G.~Genty, ``Predicting ultrafast nonlinear dynamics in fibre optics with a recurrent neural network,'' \emph{Nature machine intelligence}, vol.~3, no.~4, pp. 344--354, 2021.

\bibitem{hinton2006fast}
G.~Hinton, ``A fast learning algorithm for deep belief nets,'' \emph{Neural Computation}, 2006.

\bibitem{wan2017stacked}
X.~Wan, C.~Zhao, Y.~Wang, and W.~Liu, ``Stacked sparse autoencoder in hyperspectral data classification using spectral-spatial, higher order statistics and multifractal spectrum features,'' \emph{Infrared Physics \& Technology}, vol.~86, pp. 77--89, 2017.

\bibitem{windrim2019unsupervised}
L.~Windrim, R.~Ramakrishnan, A.~Melkumyan, R.~J. Murphy, and A.~Chlingaryan, ``Unsupervised feature-learning for hyperspectral data with autoencoders,'' \emph{Remote Sensing}, vol.~11, no.~7, p. 864, 2019.

\bibitem{singh2022enhanced}
P.~S. Singh and S.~Karthikeyan, ``Enhanced classification of remotely sensed hyperspectral images through efficient band selection using autoencoders and genetic algorithm,'' \emph{Neural Computing and Applications}, vol.~34, no.~24, pp. 21\,539--21\,550, 2022.

\bibitem{vaswani2017attention}
A.~Vaswani, N.~Shazeer, N.~Parmar, J.~Uszkoreit, L.~Jones, A.~N. Gomez, {\L}.~Kaiser, and I.~Polosukhin, ``Attention is all you need,'' \emph{Advances in neural information processing systems}, vol.~30, 2017.

\bibitem{he2021spatial}
X.~He, Y.~Chen, and Z.~Lin, ``Spatial-spectral transformer for hyperspectral image classification,'' \emph{Remote Sensing}, vol.~13, no.~3, p. 498, 2021.

\bibitem{gu2021efficiently}
A.~Gu, K.~Goel, and C.~R{\'e}, ``Efficiently modeling long sequences with structured state spaces,'' \emph{arXiv preprint arXiv:2111.00396}, 2021.

\bibitem{gu2023mamba}
A.~Gu and T.~Dao, ``Mamba: Linear-time sequence modeling with selective state spaces,'' \emph{arXiv preprint arXiv:2312.00752}, 2023.

\bibitem{kotzagiannidis2021semi}
M.~S. Kotzagiannidis and C.-B. Sch{\"o}nlieb, ``Semi-supervised superpixel-based multi-feature graph learning for hyperspectral image data,'' \emph{IEEE Transactions on Geoscience and Remote Sensing}, vol.~60, pp. 1--12, 2021.

\bibitem{chen2024prototype}
R.~Chen, H.~Yao, W.~Chen, H.~Sun, W.~Xie, L.~Dong, and X.~Lu, ``Prototype-based pseudo-label refinement for semi-supervised hyperspectral image classification,'' \emph{IEEE Geoscience and Remote Sensing Letters}, 2024.

\bibitem{tao2020semisupervised}
C.~Tao, H.~Wang, J.~Qi, and H.~Li, ``Semisupervised variational generative adversarial networks for hyperspectral image classification,'' \emph{IEEE Journal of Selected Topics in Applied Earth Observations and Remote Sensing}, vol.~13, pp. 914--927, 2020.

\bibitem{benjdira2020data}
B.~Benjdira, A.~Ammar, A.~Koubaa, and K.~Ouni, ``Data-efficient domain adaptation for semantic segmentation of aerial imagery using generative adversarial networks,'' \emph{Applied Sciences}, vol.~10, no.~3, p. 1092, 2020.

\bibitem{xue2020general}
Z.~Xue, ``A general generative adversarial capsule network for hyperspectral image spectral-spatial classification,'' \emph{Remote Sensing Letters}, vol.~11, no.~1, pp. 19--28, 2020.

\bibitem{gao2019hyperspectral}
H.~Gao, D.~Yao, M.~Wang, C.~Li, H.~Liu, Z.~Hua, and J.~Wang, ``A hyperspectral image classification method based on multi-discriminator generative adversarial networks,'' \emph{Sensors}, vol.~19, no.~15, p. 3269, 2019.

\bibitem{gan1}
H.~Li, L.~Zhang, H.~Sun, Z.~Rao, and H.~Ji, ``Discrimination of unsound wheat kernels based on deep convolutional generative adversarial network and near-infrared hyperspectral imaging technology,'' \emph{Spectrochimica Acta Part A: Molecular and Biomolecular Spectroscopy}, vol. 268, p. 120722, 2022.

\bibitem{gan2}
J.~Wang, F.~Gao, J.~Dong, and Q.~Du, ``Adaptive dropblock-enhanced generative adversarial networks for hyperspectral image classification,'' \emph{IEEE Transactions on Geoscience and Remote Sensing}, vol.~59, no.~6, pp. 5040--5053, 2020.

\bibitem{huang2019dimensionality}
H.~Huang, G.~Shi, H.~He, Y.~Duan, and F.~Luo, ``Dimensionality reduction of hyperspectral imagery based on spatial--spectral manifold learning,'' \emph{IEEE transactions on cybernetics}, vol.~50, no.~6, pp. 2604--2616, 2019.

\bibitem{li2014classification}
T.~Li, J.~Zhang, and Y.~Zhang, ``Classification of hyperspectral image based on deep belief networks,'' in \emph{2014 IEEE international conference on image processing (ICIP)}.\hskip 1em plus 0.5em minus 0.4em\relax IEEE, 2014, pp. 5132--5136.

\bibitem{johnson1998change}
R.~D. Johnson and E.~Kasischke, ``Change vector analysis: A technique for the multispectral monitoring of land cover and condition,'' \emph{International journal of remote sensing}, vol.~19, no.~3, pp. 411--426, 1998.

\bibitem{li2021unsupervised}
Q.~Li, H.~Gong, H.~Dai, C.~Li, Z.~He, W.~Wang, Y.~Feng, F.~Han, A.~Tuniyazi, H.~Li \emph{et~al.}, ``Unsupervised hyperspectral image change detection via deep learning self-generated credible labels,'' \emph{IEEE Journal of Selected Topics in Applied Earth Observations and Remote Sensing}, vol.~14, pp. 9012--9024, 2021.

\bibitem{liu2016unsupervised}
S.~Liu, L.~Bruzzone, F.~Bovolo, and P.~Du, ``Unsupervised multitemporal spectral unmixing for detecting multiple changes in hyperspectral images,'' \emph{IEEE Transactions on Geoscience and Remote Sensing}, vol.~54, no.~5, pp. 2733--2748, 2016.

\bibitem{terentev2022current}
A.~Terentev, V.~Dolzhenko, A.~Fedotov, and D.~Eremenko, ``Current state of hyperspectral remote sensing for early plant disease detection: A review,'' \emph{Sensors}, vol.~22, no.~3, p. 757, 2022.

\bibitem{mou2017unsupervised}
L.~Mou, P.~Ghamisi, and X.~X. Zhu, ``Unsupervised spectral--spatial feature learning via deep residual conv--deconv network for hyperspectral image classification,'' \emph{IEEE Transactions on Geoscience and Remote Sensing}, vol.~56, no.~1, pp. 391--406, 2017.

\bibitem{guerri2024deep}
M.~F. Guerri, C.~Distante, P.~Spagnolo, F.~Bougourzi, and A.~Taleb-Ahmed, ``Deep learning techniques for hyperspectral image analysis in agriculture: A review,'' \emph{ISPRS Open Journal of Photogrammetry and Remote Sensing}, p. 100062, 2024.

\bibitem{hu2023binary}
M.~Hu, C.~Wu, B.~Du, and L.~Zhang, ``Binary change guided hyperspectral multiclass change detection,'' \emph{IEEE Transactions on Image Processing}, vol.~32, pp. 791--806, 2023.

\bibitem{du2019unsupervised}
B.~Du, L.~Ru, C.~Wu, and L.~Zhang, ``Unsupervised deep slow feature analysis for change detection in multi-temporal remote sensing images,'' \emph{IEEE Transactions on Geoscience and Remote Sensing}, vol.~57, no.~12, pp. 9976--9992, 2019.

\bibitem{hu2022hypernet}
M.~Hu, C.~Wu, and L.~Zhang, ``Hypernet: Self-supervised hyperspectral spatial--spectral feature understanding network for hyperspectral change detection,'' \emph{IEEE Transactions on Geoscience and Remote Sensing}, vol.~60, pp. 1--17, 2022.

\bibitem{simon2022convolutional}
A.-M. Sim{\'o}n~S{\'a}nchez, J.~Gonz{\'a}lez-Piqueras, L.~de~la Ossa, and A.~Calera, ``Convolutional neural networks for agricultural land use classification from sentinel-2 image time series,'' \emph{Remote Sensing}, vol.~14, no.~21, p. 5373, 2022.

\bibitem{gong2019novel}
Z.~Gong, P.~Zhong, W.~Hu, Z.~Xiao, and X.~Yin, ``A novel statistical metric learning for hyperspectral image classification,'' \emph{arXiv preprint arXiv:1905.05087}, 2019.

\bibitem{zhan2017semisupervised}
Y.~Zhan, D.~Hu, Y.~Wang, and X.~Yu, ``Semisupervised hyperspectral image classification based on generative adversarial networks,'' \emph{IEEE Geoscience and Remote Sensing Letters}, vol.~15, no.~2, pp. 212--216, 2017.

\bibitem{madhavan2023wheat}
J.~Madhavan, M.~Salim, U.~Durairaj, and R.~Kotteeswaran, ``Wheat seed classification using neural network pattern recognizer,'' \emph{Materials Today: Proceedings}, vol.~81, pp. 341--345, 2023.

\bibitem{sethy2022identification}
P.~K. Sethy, ``Identification of wheat tiller based on alexnet-feature fusion,'' \emph{Multimedia Tools and Applications}, vol.~81, no.~6, pp. 8309--8316, 2022.

\bibitem{site4}
``Wheat kernels,'' \url{https://archive.ics.uci.edu/datasets?search=wheat+}, 2023.

\bibitem{dreier2022hyperspectral}
E.~S. Dreier, K.~M. Sorensen, T.~Lund-Hansen, B.~M. Jespersen, and K.~S. Pedersen, ``Hyperspectral imaging for classification of bulk grain samples with deep convolutional neural networks,'' \emph{Journal of Near Infrared Spectroscopy}, vol.~30, no.~3, pp. 107--121, 2022.

\bibitem{lingwal2021image}
S.~Lingwal, K.~K. Bhatia, and M.~S. Tomer, ``Image-based wheat grain classification using convolutional neural network,'' \emph{Multimedia Tools and Applications}, pp. 1--25, 2021.

\bibitem{sabanci2017computer}
K.~Sabanci, A.~Kayabasi, and A.~Toktas, ``Computer vision-based method for classification of wheat grains using artificial neural network,'' \emph{Journal of the Science of Food and Agriculture}, vol.~97, no.~8, pp. 2588--2593, 2017.

\bibitem{hu2015deep}
W.~Hu, Y.~Huang, L.~Wei, F.~Zhang, and H.~Li, ``Deep convolutional neural networks for hyperspectral image classification,'' \emph{Journal of Sensors}, vol. 2015, no.~1, p. 258619, 2015.

\bibitem{slavkovikj2015hyperspectral}
V.~Slavkovikj, S.~Verstockt, W.~De~Neve, S.~Van~Hoecke, and R.~Van~de Walle, ``Hyperspectral image classification with convolutional neural networks,'' in \emph{Proceedings of the 23rd ACM international conference on Multimedia}, 2015, pp. 1159--1162.

\bibitem{david2020global}
E.~David, S.~Madec, P.~Sadeghi-Tehran, H.~Aasen, B.~Zheng, S.~Liu, N.~Kirchgessner, G.~Ishikawa, K.~Nagasawa, M.~A. Badhon \emph{et~al.}, ``Global wheat head detection (gwhd) dataset: a large and diverse dataset of high-resolution rgb-labelled images to develop and benchmark wheat head detection methods,'' \emph{Plant Phenomics}, 2020.

\bibitem{caporaso2018near}
N.~Caporaso, M.~B. Whitworth, and I.~D. Fisk, ``Near-infrared spectroscopy and hyperspectral imaging for non-destructive quality assessment of cereal grains,'' \emph{Applied spectroscopy reviews}, vol.~53, no.~8, pp. 667--687, 2018.

\bibitem{bao2019rapid}
Y.~Bao, C.~Mi, N.~Wu, F.~Liu, and Y.~He, ``Rapid classification of wheat grain varieties using hyperspectral imaging and chemometrics,'' \emph{Applied Sciences}, vol.~9, no.~19, p. 4119, 2019.

\bibitem{ozkan2021wheat}
K.~{\"O}zkan, S.~Erol, and I.~{\c{S}}ahin, ``Wheat kernels classification using visible-near infrared camera based on deep learning,'' \emph{Pamukkale {\"U}niversitesi M{\"u}hendislik Bilimleri Dergisi}, vol.~27, no.~5, pp. 618--626, 2021.

\bibitem{yipeng2022determination}
L.~Yipeng, L.~Wenbing, H.~Kaixuan, T.~Wentao, Z.~Ling, W.~Shizhuang, and H.~Linsheng, ``Determination of wheat kernels damaged by fusarium head blight using monochromatic images of effective wavelengths from hyperspectral imaging coupled with an architecture self-search deep network,'' \emph{Food Control}, vol. 135, p. 108819, 2022.

\bibitem{zhu2023identification}
J.~Zhu, H.~Li, Z.~Rao, and H.~Ji, ``Identification of slightly sprouted wheat kernels using hyperspectral imaging technology and different deep convolutional neural networks,'' \emph{Food Control}, vol. 143, p. 109291, 2023.

\bibitem{que2023identification}
H.~Que, X.~Zhao, X.~Sun, Q.~Zhu, and M.~Huang, ``Identification of wheat kernel varieties based on hyperspectral imaging technology and grouped convolutional neural network with feature intervals,'' \emph{Infrared Physics \& Technology}, vol. 131, p. 104653, 2023.

\bibitem{wan2022hyperspectral}
L.~Wan, H.~Li, C.~Li, A.~Wang, Y.~Yang, and P.~Wang, ``Hyperspectral sensing of plant diseases: Principle and methods,'' \emph{Agronomy}, vol.~12, no.~6, p. 1451, 2022.

\bibitem{terentev2023hyperspectral}
A.~Terentev, V.~Badenko, E.~Shaydayuk, D.~Emelyanov, D.~Eremenko, D.~Klabukov, A.~Fedotov, and V.~Dolzhenko, ``Hyperspectral remote sensing for early detection of wheat leaf rust caused by puccinia triticina,'' \emph{Agriculture}, vol.~13, no.~6, p. 1186, 2023.

\bibitem{feng2024wheat}
G.~Feng, Y.~Gu, C.~Wang, Y.~Zhou, S.~Huang, and B.~Luo, ``Wheat fusarium head blight automatic non-destructive detection based on multi-scale imaging: A technical perspective,'' \emph{Plants}, vol.~13, no.~13, p. 1722, 2024.

\bibitem{jin2018classifying}
X.~Jin, L.~Jie, S.~Wang, H.~J. Qi, and S.~W. Li, ``Classifying wheat hyperspectral pixels of healthy heads and fusarium head blight disease using a deep neural network in the wild field,'' \emph{Remote Sensing}, vol.~10, no.~3, p. 395, 2018.

\bibitem{khotimah2023mce}
W.~N. Khotimah, M.~Bennamoun, F.~Boussaid, L.~Xu, D.~Edwards, and F.~Sohel, ``Mce-st: Classifying crop stress using hyperspectral data with a multiscale conformer encoder and spectral-based tokens,'' \emph{International Journal of Applied Earth Observation and Geoinformation}, vol. 118, p. 103286, 2023.

\bibitem{khan2021early}
I.~H. Khan, H.~Liu, W.~Li, A.~Cao, X.~Wang, H.~Liu, T.~Cheng, Y.~Tian, Y.~Zhu, W.~Cao \emph{et~al.}, ``Early detection of powdery mildew disease and accurate quantification of its severity using hyperspectral images in wheat,'' \emph{Remote Sensing}, vol.~13, no.~18, p. 3612, 2021.

\bibitem{almoujahed2022detection}
M.~B. Almoujahed, A.~K. Rangarajan, R.~L. Whetton, D.~Vincke, D.~Eylenbosch, P.~Vermeulen, and A.~M. Mouazen, ``Detection of fusarium head blight in wheat under field conditions using a hyperspectral camera and machine learning,'' \emph{Computers and Electronics in Agriculture}, vol. 203, p. 107456, 2022.

\bibitem{rangarajan2022detection}
A.~K. Rangarajan, R.~L. Whetton, and A.~M. Mouazen, ``Detection of fusarium head blight in wheat using hyperspectral data and deep learning,'' \emph{Expert Systems with Applications}, vol. 208, p. 118240, 2022.

\bibitem{feng2022hyperspectral}
Z.-H. Feng, L.-Y. Wang, Z.-Q. Yang, Y.-Y. Zhang, X.~Li, L.~Song, L.~He, J.-Z. Duan, and W.~Feng, ``Hyperspectral monitoring of powdery mildew disease severity in wheat based on machine learning,'' \emph{Frontiers in Plant Science}, vol.~13, p. 828454, 2022.

\bibitem{xu2023wheat}
L.~Xu, B.~Cao, F.~Zhao, S.~Ning, P.~Xu, W.~Zhang, and X.~Hou, ``Wheat leaf disease identification based on deep learning algorithms,'' \emph{Physiological and Molecular Plant Pathology}, vol. 123, p. 101940, 2023.

\bibitem{sereda2023development}
I.~Sereda, R.~Danilov, O.~Kremneva, M.~Zimin, and Y.~Podushin, ``Development of methods for remote monitoring of leaf diseases in wheat agrocenoses,'' \emph{Plants}, vol.~12, no.~18, p. 3223, 2023.

\bibitem{blasch2023potential}
G.~Blasch, T.~Anberbir, T.~Negash, L.~Tilahun, F.~Y. Belayineh, Y.~Alemayehu, G.~Mamo, D.~P. Hodson, and F.~A. Rodrigues~Jr, ``The potential of uav and very high-resolution satellite imagery for yellow and stem rust detection and phenotyping in ethiopia,'' \emph{Scientific Reports}, vol.~13, no.~1, p. 16768, 2023.

\bibitem{chang2024recognition}
S.~Chang, G.~Yang, J.~Cheng, Z.~Feng, Z.~Fan, X.~Ma, Y.~Li, X.~Yang, and C.~Zhao, ``Recognition of wheat rusts in a field environment based on improved densenet,'' \emph{Biosystems Engineering}, vol. 238, pp. 10--21, 2024.

\bibitem{ma2022applications}
J.~Ma, B.~Zheng, and Y.~He, ``Applications of a hyperspectral imaging system used to estimate wheat grain protein: A review,'' \emph{Frontiers in Plant Science}, vol.~13, p. 837200, 2022.

\bibitem{barbedo2023review}
J.~G.~A. Barbedo, ``A review on the combination of deep learning techniques with proximal hyperspectral images in agriculture,'' \emph{Computers and Electronics in Agriculture}, vol. 210, p. 107920, 2023.

\bibitem{hu2021predicting}
N.~Hu, W.~Li, C.~Du, Z.~Zhang, Y.~Gao, Z.~Sun, L.~Yang, K.~Yu, Y.~Zhang, and Z.~Wang, ``Predicting micronutrients of wheat using hyperspectral imaging,'' \emph{Food Chemistry}, vol. 343, p. 128473, 2021.

\bibitem{song2021assessment}
Y.~Song, G.~Teng, Y.~Yuan, T.~Liu, and Z.~Sun, ``Assessment of wheat chlorophyll content by the multiple linear regression of leaf image features,'' \emph{Information processing in Agriculture}, vol.~8, no.~2, pp. 232--243, 2021.

\bibitem{yang2021estimation}
B.~Yang, J.~Ma, X.~Yao, W.~Cao, and Y.~Zhu, ``Estimation of leaf nitrogen content in wheat based on fusion of spectral features and deep features from near infrared hyperspectral imagery,'' \emph{Sensors}, vol.~21, no.~2, p. 613, 2021.

\bibitem{astaoui2021mapping}
G.~Astaoui, J.~E. Dadaiss, I.~Sebari, S.~Benmansour, and E.~Mohamed, ``Mapping wheat dry matter and nitrogen content dynamics and estimation of wheat yield using uav multispectral imagery machine learning and a variety-based approach: Case study of morocco,'' \emph{AgriEngineering}, vol.~3, no.~1, pp. 29--49, 2021.

\bibitem{lu2022assessment}
N.~Lu, Y.~Wu, H.~Zheng, X.~Yao, Y.~Zhu, W.~Cao, and T.~Cheng, ``An assessment of multi-view spectral information from uav-based color-infrared images for improved estimation of nitrogen nutrition status in winter wheat,'' \emph{Precision Agriculture}, vol.~23, no.~5, pp. 1653--1674, 2022.

\bibitem{du2023incremental}
R.~Du, J.~Chen, Y.~Xiang, Z.~Zhang, N.~Yang, X.~Yang, Z.~Tang, H.~Wang, X.~Wang, H.~Shi \emph{et~al.}, ``Incremental learning for crop growth parameters estimation and nitrogen diagnosis from hyperspectral data,'' \emph{Computers and Electronics in Agriculture}, vol. 215, p. 108356, 2023.

\bibitem{cheng2022wheat}
E.~Cheng, B.~Zhang, D.~Peng, L.~Zhong, L.~Yu, Y.~Liu, C.~Xiao, C.~Li, X.~Li, Y.~Chen \emph{et~al.}, ``Wheat yield estimation using remote sensing data based on machine learning approaches,'' \emph{Frontiers in Plant Science}, vol.~13, p. 1090970, 2022.

\bibitem{moghimi2020aerial}
A.~Moghimi, C.~Yang, and J.~A. Anderson, ``Aerial hyperspectral imagery and deep neural networks for high-throughput yield phenotyping in wheat,'' \emph{Computers and Electronics in Agriculture}, vol. 172, p. 105299, 2020.

\bibitem{liu2023estimation}
Y.~Liu, L.~Sun, B.~Liu, Y.~Wu, J.~Ma, W.~Zhang, B.~Wang, and Z.~Chen, ``Estimation of winter wheat yield using multiple temporal vegetation indices derived from uav-based multispectral and hyperspectral imagery,'' \emph{Remote Sensing}, vol.~15, no.~19, p. 4800, 2023.

\bibitem{li2022toward}
K.-Y. Li, R.~Sampaio~de Lima, N.~G. Burnside, E.~Vahtm{\"a}e, T.~Kutser, K.~Sepp, V.~H. Cabral~Pinheiro, M.-D. Yang, A.~Vain, and K.~Sepp, ``Toward automated machine learning-based hyperspectral image analysis in crop yield and biomass estimation,'' \emph{Remote Sensing}, vol.~14, no.~5, p. 1114, 2022.

\bibitem{fei2022application}
S.~Fei, M.~A. Hassan, Y.~Xiao, A.~Rasheed, X.~Xia, Y.~Ma, L.~Fu, Z.~Chen, and Z.~He, ``Application of multi-layer neural network and hyperspectral reflectance in genome-wide association study for grain yield in bread wheat,'' \emph{Field Crops Research}, vol. 289, p. 108730, 2022.

\bibitem{yang2023rapid}
Y.~Yang, R.~Nan, T.~Mi, Y.~Song, F.~Shi, X.~Liu, Y.~Wang, F.~Sun, Y.~Xi, and C.~Zhang, ``Rapid and nondestructive evaluation of wheat chlorophyll under drought stress using hyperspectral imaging,'' \emph{International Journal of Molecular Sciences}, vol.~24, no.~6, p. 5825, 2023.

\bibitem{zhang2024comparison}
S.~Zhang, X.~Qi, J.~Duan, X.~Yuan, H.~Zhang, W.~Feng, T.~Guo, and L.~He, ``Comparison of attention mechanism-based deep learning and transfer strategies for wheat yield estimation using multisource temporal drone imagery,'' \emph{IEEE Transactions on Geoscience and Remote Sensing}, vol.~62, pp. 1--23, 2024.

\bibitem{zhao2024spatial}
R.~Zhao, W.~Tang, M.~Liu, N.~Wang, H.~Sun, M.~Li, and Y.~Ma, ``Spatial-spectral feature extraction for in-field chlorophyll content estimation using hyperspectral imaging,'' \emph{Biosystems Engineering}, vol. 246, pp. 263--276, 2024.

\bibitem{liu2024estimation}
T.~Liu, T.~Yang, S.~Zhu, N.~Mou, W.~Zhang, W.~Wu, Y.~Zhao, Z.~Yao, J.~Sun, C.~Chen \emph{et~al.}, ``Estimation of wheat biomass based on phenological identification and spectral response,'' \emph{Computers and Electronics in Agriculture}, vol. 222, p. 109076, 2024.

\end{thebibliography}

\end{document}